\pdfoutput=1
\documentclass[10pt,twocolumn,letterpaper]{article}

\usepackage[pagenumbers]{pprint}

\usepackage{cuted}
\usepackage{currfile}
\usepackage{caption}
\usepackage{subcaption}
\usepackage{booktabs}
\usepackage{multirow}
\usepackage{colortbl}
\usepackage{xspace}
\usepackage{amsmath,amssymb,bm}
\usepackage{graphicx}
\usepackage{tikz}
\usetikzlibrary{arrows.meta,positioning,calc,fit,backgrounds,patterns}
\usepackage{enumitem}
\setlist{nosep,leftmargin=*}
\usepackage{algorithm}
\usepackage[noend]{algpseudocode}
\usepackage{pifont}
\newcommand{\cmark}{\ding{51}}
\newcommand{\xmark}{\ding{55}}
\algrenewcommand\algorithmiccomment[1]{\hfill{\color{gray}\(\triangleright\) #1}}

\newcommand{\method}{PlayRender\xspace}

\definecolor{prRef}{RGB}{206,186,230}      \definecolor{prRefLine}{RGB}{134,104,188}
\definecolor{prFirst}{RGB}{255,185,115}    \definecolor{prFirstLine}{RGB}{222,134,46}
\definecolor{prCtl}{RGB}{64,64,64}
\definecolor{prGen}{RGB}{184,210,232}      \definecolor{prGenLine}{rgb}{0.21,0.49,0.74}
\definecolor{prFail}{RGB}{214,90,60}
\newcommand{\rowlab}[2][]{\parbox[c]{1.55cm}{\raggedleft\footnotesize #2\ifx\relax#1\relax\else\\[-2pt]{\scriptsize\color{black!55}#1}\fi}}
\newcommand{\ours}{\textbf{\color{prGenLine}Ours}}

\renewcommand{\paragraph}[1]{\vspace{.4em}\noindent\textbf{#1.}}

\definecolor{linkblue}{rgb}{0.21,0.49,0.74}
\usepackage[pagebackref,breaklinks,colorlinks,allcolors=linkblue]{hyperref}
\hypersetup{pdftitle={PlayRender: Streaming Neural Rendering for Interactive Worlds},pdfauthor={Ruiqi Wang}}

\title{\method: Streaming Neural Rendering\\for Interactive Worlds}

\author{Ruiqi Wang\\
Department of Physics, University of California San Diego\\
La Jolla, CA 92093, USA\\
{\tt\small riw010@ucsd.edu}
}

\begin{document}
\maketitle
\begin{strip}
\centering
\begin{tikzpicture}[x=1cm, y=1cm, font=\scriptsize]
\def\TW{2.47}\def\TH{1.389375}\def\TG{0.05}
\pgfmathsetmacro{\RH}{(\TH-\TG)/2}\pgfmathsetmacro{\RW}{\RH*16/9}
\pgfmathsetmacro{\xF}{\RW+0.12}\pgfmathsetmacro{\xG}{\xF+\TW+0.12}
\pgfmathsetmacro{\xE}{\xG+3*\TW+2*\TG}\pgfmathsetmacro{\GH}{3*\TH+2*\TG}
\foreach \c/\r in {n1/0,n2/1,n3/2} {
  \pgfmathsetmacro{\y}{\GH-\TH-\r*(\TH+\TG)}
  \node[inner sep=0, anchor=south west] at (0,\y+\RH+\TG) {\includegraphics[width=\RW cm]{fig/img/tz_\c_chars.jpg}};
  \node[inner sep=0, anchor=south west] at (0,\y) {\includegraphics[width=\RW cm]{fig/img/tz_\c_scene.jpg}};
  \node[inner sep=0, anchor=south west] at (\xF,\y) {\includegraphics[width=\TW cm]{fig/img/tz_\c_0.jpg}};
  \foreach \k in {1,2,3} {
    \pgfmathsetmacro{\x}{\xG+(\k-1)*(\TW+\TG)}
    \node[inner sep=0, anchor=south west] at (\x,\y) {\includegraphics[width=\TW cm]{fig/img/tz_\c_\k.jpg}};
  }
}
\pgfmathsetmacro{\yr}{\GH+0.075}\pgfmathsetmacro{\yl}{\GH+0.13}
\draw[prRefLine, line width=1.1pt] (0,\yr) -- (\RW,\yr);
\draw[prFirstLine, line width=1.1pt] (\xF,\yr) -- (\xF+\TW,\yr);
\draw[prGenLine, line width=1.1pt, -{Stealth[length=1.6mm, width=1.4mm]}] (\xG,\yr) -- (\xE,\yr);
\pgfmathsetmacro{\yb}{\GH+0.25}%
\node[anchor=base, inner sep=0pt] at (\RW/2,\yb) {References};
\node[anchor=base, inner sep=0pt] at (\xF+\TW/2,\yb) {First frame (given)};
\node[anchor=base, inner sep=0pt] at (\xG+\TW/2,\yb) {Generated, $1.5$\,s};
\node[anchor=base, inner sep=0pt] at (\xG+1.5*\TW+\TG,\yb) {$3.0$\,s};
\node[anchor=base, inner sep=0pt] at (\xG+2.5*\TW+2*\TG,\yb) {$4.5$\,s};
\node[anchor=south west, inner sep=1.2pt, fill=white, text=black, font=\fontsize{5.6}{6}\selectfont] at (\xG,\GH-\TH+0.475) {control};
\def\CW{5.55}\pgfmathsetmacro{\xC}{17.4-\CW}
\node[inner sep=0, anchor=south west] at (\xC,-0.02) {\includegraphics[width=\CW cm]{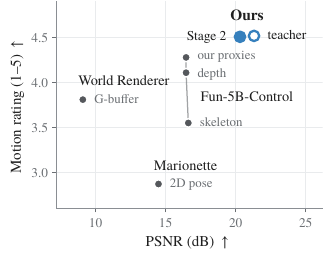}};
\draw[black!45, line width=1.1pt] (\xC+0.172*\CW,\yr) -- (\xC+0.984*\CW,\yr);
\node[anchor=base, inner sep=0pt] at (\xC+0.578*\CW,\yb) {Comparison on $66$ engine renders};
\end{tikzpicture}
\captionsetup{hypcap=false}
\captionof{figure}{\method turns a render of untextured 3D proxies from a game engine (insets), a first frame and reference images into game video, chunk by chunk. Left: three test clips in three environments; none of their characters was seen in training. Right: our causal student (Stage~2, four steps) and its offline teacher ($50$ steps) against released renderers (gray), labeled by control signal.}
\captionsetup{hypcap=true}
\label{fig:teaser}
\end{strip}

\begin{abstract}
Rendering a game simulation into game-quality video with a generative model offers a path toward neural rendering for interactive worlds, provided that the video follows the simulation precisely and is produced as the simulation unfolds.
We introduce \method, a streaming neural renderer behind a game engine: the engine rasterizes its untextured 3D proxies and thereby determines what happens, including motion and contacts between characters, and a first frame and reference images of the characters and the scene determine how it looks.
A bidirectional teacher learns this control by packing control and target latents into the same frames of a video diffusion transformer, and we distill it into a block-causal four-step student: because control and target share latent frames, one mask makes both causal, so the student generates chunk by chunk from past and present control only, with the references kept as an attention sink.
We train on engine renders of one game state with proxies and with detailed assets and on gameplay and external videos with reconstructed proxies.
On held-out engine renders of two unseen characters, the student stays within $1.0$\,dB PSNR of its teacher at $11.7{\times}$ less time, and a change of control takes effect in the first frame of the next chunk; eight further characters, rendered after all models were fixed, cost $0.8$\,dB.
Over one-minute rollouts the references in its cache reduce appearance drift, and it renders $4.5$ frames per second on one A100 and $17$, above real time, with one denoising step on three, where a control switch shows after $3.7$\,s.
Released renderers of the same size, not trained on our data, stay behind it on each of the $66$ engine clips we compared, by $3.7$\,dB on average, and not only in image quality: where the silhouettes of two characters meet, a pixel measure finds the moving weapon or limb at its frame and place in $68\%$ of the measurable events for our student and in $4$ to $14\%$ for them.
Where it is a weapon, ten volunteers, blind to the method, find it missing in $16\%$ of their answers for our student and in $51$ and $65\%$ for a released model driven by our proxies or by skeletons, although our student shows it fully in only $31\%$ (ground truth: $95\%$).
Trained under one reduced budget, a teacher with skeletons instead of proxies reaches the same PSNR but loses the weapon in $18\%$ of the ratings against $2\%$ and passes this measure in $30\%$ of the events against $73\%$.
\end{abstract}

\section{Introduction}
\label{sec:intro}

An interactive world must respond to decisions, preserve the identities of its inhabitants, and show the consequences of their actions.
Generative models make visually rich simulation plausible, from learned interactive environments~\cite{genie} to diffusion-based game engines~\cite{diamond,gamengen} and streaming video world models~\cite{matrixgame2,gamecraft,yume15,solarwm}.
Yet plausibility is not enough for a playable encounter: when a player strikes an opponent, both bodies and their weapons must be where the game puts them, and the contact must occur at the right time.
A weapon that is missing at the strike is not a loss of image quality but a wrong frame.

A practical route is to separate programmable simulation from appearance generation.
A lightweight engine with untextured assets, collision rules, articulated motion and a controllable camera defines the game, and a generative renderer turns its inexpensive output into a rich appearance; the scene state stays inspectable and editable.
Recent renderers follow this route with G-buffers~\cite{framediffuser,genworldrenderer}, boxes~\cite{programmableworld}, poses or skeletons~\cite{marionette,wizard} or untextured frames~\cite{magpie,anygamerenderer}.
We study the rendering component of this route, with an emphasis on multi-character control and the constraints of streaming deployment; learning game rules is outside our scope.

\noindent\textbf{What the engine hands over.}
Pose sequences, masks and spatial guidance have enabled substantial progress in controllable video synthesis and character animation~\cite{controlnet,animateanyone,vace,wananimate,scail}, and they are convenient to recover from existing videos.
A game engine has no need to estimate them: it holds articulated bodies, props and the camera in 3D and can rasterize its untextured proxies with depth testing, in the image coordinates of the desired output.
Such a render contains weapons and other props, which skeletons omit (\cref{sec:representation}), and none of the albedo and material channels through which G-buffer renderers~\cite{framediffuser,genworldrenderer} tie the appearance to finished assets.
We adopt it as the control and take the appearance from a first frame and reference images.

Following this control in a stream raises three challenges.
\emph{First, the control must be dense and causal.}
We condition in context: clean control tokens and noisy target tokens are packed into the same latent frames and share every attention layer, so the target sees the full control grid without an adapter and a single block-causal mask makes control and output causal together.
\emph{Second, the renderer must respond within a practical compute budget.}
Bidirectional generation over a full window depends on future control and on many denoising steps.
We build on a 5B video backbone, convert a bidirectional teacher into a few-step causal student~\cite{causvid,selfforcing,solarwm}, and measure throughput, time to first frame and the latency of a control switch separately.
\emph{Third, control and appearance must persist under generated history.}
A model trained only on clean history meets its own errors at inference, and they propagate across chunks~\cite{diffusionforcing,selfforcing}.
We train the student on its own rollouts with a bounded history, keep the references in its cache as an attention sink, and measure quality after one minute separately from short-clip reconstruction.

We introduce \method, a streaming neural renderer that turns untextured 3D proxies from a game engine, a first frame and reference images into game video, generated chunk by chunk from past and present control (\cref{fig:teaser,fig:overview}).
A bidirectional teacher learns joint control--target attention, and the two stages of SolarWM~\cite{solarwm,anyflow} convert it into a four-step block-causal student whose control is causal as well, so the engine provides control only one chunk of $0.75$\,s ahead.
Training pairs come from engine renders that drive proxies and detailed assets with one game state, and from gameplay footage of five commercial games and external videos with proxies reconstructed from the video.
On engine renders of two unseen characters, the student stays within $1.0$\,dB PSNR of its teacher at $11.7{\times}$ less time (\cref{tab:main}), a change of control takes effect in the first frame of the next chunk, and eight further characters, rendered after all models were fixed, cost $0.8$\,dB.
It sustains one-minute rollouts, in which its references reduce appearance drift, and renders $4.5$ frames per second on one A100 and $17$ with one denoising step on three (switch latency $3.7$\,s).
Released renderers of the same size, not trained on our data, stay behind it on each of the $66$ engine clips we compared, by $3.7$\,dB on average, and rarely show a strike at its frame and place (in $4$ to $14\%$ of the contact events, against $68\%$); in a blind human check, one of them with our proxies or skeletons loses a striking weapon in $51$ and $65\%$ of the answers, our student in $16\%$, but it shows the weapon fully in only $31\%$ (ground truth: $95\%$).
On external videos, our students lose fidelity.

Our contributions are:
(1) a streaming renderer for multi-character interaction behind a game engine: in-context conditioning of a 5B video model in which control and target share latent frames, so that the block-causal mask of the student applies to both without change, distilled into a few-step student that keeps its references as an attention sink;
(2) evidence that the output follows the control under swaps and switches, and that full-frame fidelity hides what happens at contacts, from a matched comparison of proxies and skeletons, a pixel measure and a blind human check;
(3) paired data from engine renders, gameplay and external videos, and measurements of one-minute stability, time to first frame and switch latency.

\section{Related Work}
\label{sec:related}

\paragraph{Generative game engines}
Generative models have been used to replace game engines entirely~\cite{gamengen,diamond,genie,thematrix} and to generate and control open-world game videos~\cite{gamegenx,gamefactory,matrixgame,matrixgame2,matrixgame4,gamecraft,yan,yume15,solarwm} and two-player fights from keyboard input in real time~\cite{wantofight}.
Game world models learn rules and physics from pixels and are steered through actions, language, predicted game states or the decisions of a rule engine~\cite{reactivegwm,incantation,worldmind,stateplay,gamedirector}; in all of them, the generator still decides how characters move and interact.
A recent review~\cite{luo2026ai} covers AI for games in the era of foundation models, from agents that play to learned game worlds and generation at runtime.

\paragraph{Neural rendering behind an engine}
A second line keeps the engine and replaces its final rendering.
Image translation maps semantic labels or rendered frames to realistic images~\cite{vid2vid,epe}, and neural renderers map G-buffers with albedo and material channels to frames~\cite{framediffuser,genworldrenderer,genworldrendererfast,neuralsrender}, so the appearance stays tied to the engine assets; video models are also conditioned on buffers of animated meshes with a reference image~\cite{nativerenderer4d} and on oriented boxes from a programmable engine~\cite{programmableworld}.
Marionette~\cite{marionette} rasterizes predicted character states into a pose-control video and paints it from a first frame, and SolarRenderer~\cite{solarrenderer} renders structure videos, including untextured scene layouts, with an optional reference image.
Closest to our setting, Magpie~\cite{magpie} renders untextured engine frames from a first frame in real time for exploration and driving, Wizard~\cite{wizard} renders scene geometry with posed skeletons and per-character reference images with a causal student, and AnyGameRenderer~\cite{anygamerenderer} renders blockout videos of many entities with a 33B model and scores their counts and timing.
We focus on close combat, where a weapon or limb passes the opponent within a few frames: our control renders every character and prop as a depth-tested proxy, which boxes do not articulate and skeletons omit (\cref{sec:representation}), and we measure whether the part appears at its frame and place.

\paragraph{Reference-driven character animation}
Animating a reference image with a driving signal has progressed from UNet-based models with reference networks and pose guiders~\cite{animateanyone,magicanimate,champ,mimicmotion,unianimate,animatex,stableanimator,mimo} to video DiTs~\cite{humandit,unianimatedit,vace,dreamactorm1,wananimate,dynamictrl}, driven mostly by 2D skeletons~\cite{dwpose}, sometimes by rendered bodies or 3D-consistent poses~\cite{champ,scail}, also for several characters with one reference each~\cite{multianimate,instanceanimator,magicfight}; they animate bidirectionally from skeletons or masks, whereas our control also carries props and we generate causally.

\paragraph{Causal and autoregressive video diffusion}
Diffusion Forcing~\cite{diffusionforcing} trains with independent per-frame noise levels, CausVid~\cite{causvid} distills a bidirectional teacher into a few-step causal student, and Self Forcing~\cite{selfforcing} trains on the student's own autoregressive rollouts, a line extended by~\cite{selfforcingpp,causalforcing,apt2,selfgradientforcing}.
Streaming models push length and latency~\cite{magi1,skyreelsv2,framepack,pyramidflow,streamdit,streamingt2v,fifodiffusion,longlive,rollingforcing}, and bounded caches with sinks or re-aligned positions extend sequences beyond the training length~\cite{streamingllm,infinityrope,deepforcing}.
MotionStream~\cite{motionstream} and LiveAnimate~\cite{liveanimate} stream video from motion tracks and from the pose of one person.
SolarWM~\cite{solarwm} combines a bidirectional stage, a teacher-forced initialization with the flow maps of AnyFlow~\cite{anyflow} and distribution matching for camera-controlled world models; our student follows its two stages, with the camera branch replaced by dense in-context control.

\section{Method}
\label{sec:method}

\begin{figure*}[t]
\centering
\input{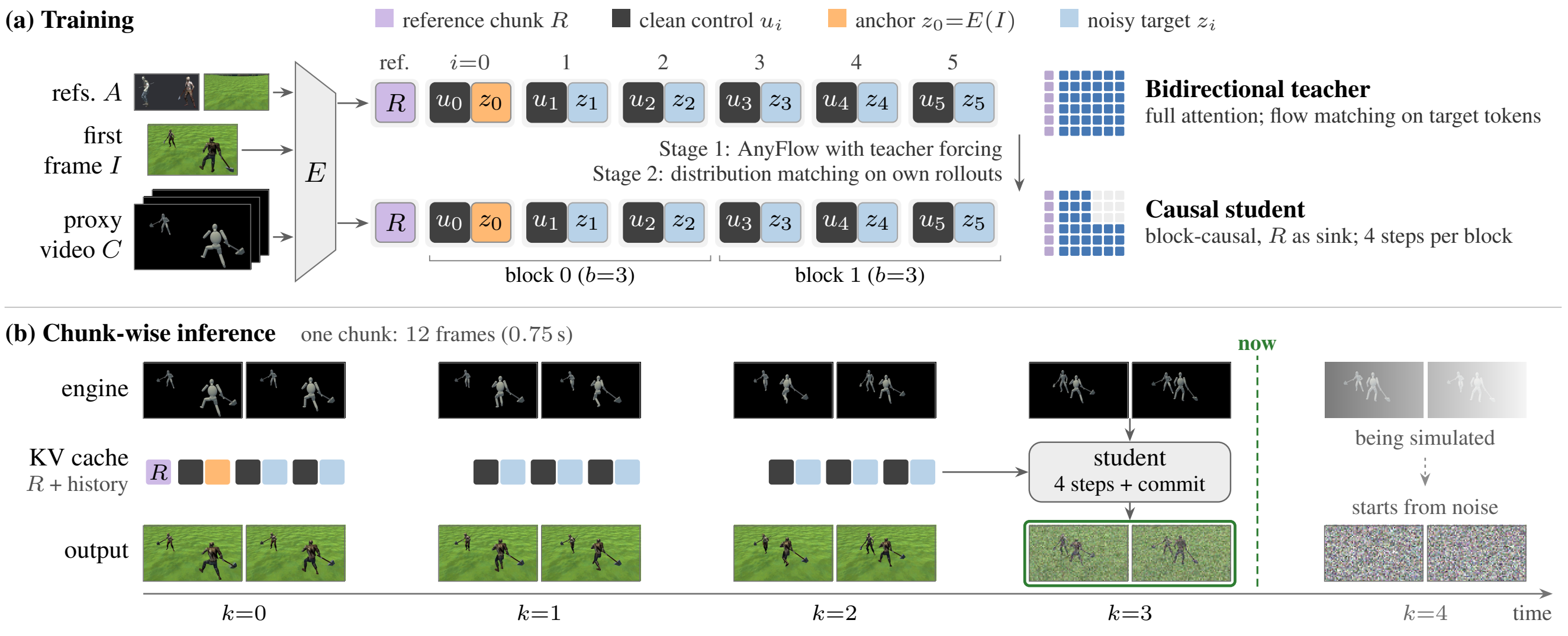}
\caption{\textbf{Overview of \method.}
(a) Each latent frame packs the clean control $u_i$ and the target $z_i$ side by side, after the reference chunk $R$.
The teacher attends over all packed frames; the student uses a block-causal mask, so block $k$ sees control only up to block $k$, and $R$ stays visible.
(b) The engine provides the control of one chunk ($0.75$\,s) at a time; the student denoises it in four steps from its control, the reference sink and up to $h{=}15$ latent frames of history, and commits it to the KV cache without reading any later control.}
\label{fig:overview}
\end{figure*}

Let $s_t$ denote a game state with articulated characters, their transforms, props and the camera.
A simulator updates $s_{t+1}=F(s_t,a_t)$ under an input $a_t$, and a lightweight rasterizer produces a control image $c_t=\mathcal{R}_{\text{low}}(s_t)$ in which the characters and their props are untextured 3D proxies on a black background.
Given a control sequence $C=(c_0,\dots,c_{T-1})$, appearance inputs and an optional text description $q$, we model
\begin{equation}
p_\theta(Y\mid C,I,A,q),\qquad y_0=I \text{ if } I \text{ is given}.
\label{eq:task}
\end{equation}
The appearance inputs, each optional, are a first frame $I$ of the starting scene and references $A=(a_{\text{c}},a_{\text{s}})$, an image of the characters and one of the scene: the engine determines what happens, and $I$ and $A$ determine how it looks.
We aim at control along three axes:
\begin{enumerate}[label=\textbf{(R\arabic*)}, leftmargin=*, nosep]
\item \emph{Motion control}: the generated characters, props and camera follow the proxies, including contacts and occlusions between characters.
\item \emph{Appearance control}: characters and scene keep the appearance given by $I$ and $A$, also in long sequences.
\item \emph{Causal control}: every output frame depends only on the appearance inputs and on control that the engine has already produced, so control is needed only a short, fixed time ahead and a change of control takes effect after a bounded delay.
\end{enumerate}
We call generation under R3, with control read and video emitted one chunk at a time, \emph{streaming} rendering; the text $q$ is fixed for a rollout and must then not describe future control (\cref{sec:ablation}).
A bidirectional teacher learns R1 and R2 by packing control and target into the same latent frames (\cref{sec:teacher}), and the two stages of SolarWM~\cite{solarwm} convert it into a block-causal student for R3 (\cref{sec:student}, \cref{fig:overview}).

\subsection{Backbone and representations}
\label{sec:prelim}

We build on Wan2.2-TI2V-5B~\cite{wan2025} with its video VAE $E$ and text encoder, and follow the training recipe of SolarWM~\cite{solarwm}.
Videos are processed at $480{\times}864$ and $16$\,fps.
An $81$-frame window of $5$\,s maps to $21$ latent frames of shape $48{\times}30{\times}54$, where the first latent encodes the first frame alone and every further latent encodes four frames.
Target video and control are encoded into latents $z=E(Y)$ and $u=E(C)$, and each reference image is encoded as a single latent frame.
When the first frame is given, $z_0=E(I)$ is a clean anchor that stays fixed throughout denoising and is excluded from all losses.
The control is rendered with the game camera and depth testing, so every control pixel shows the visible surface of a body part or prop, and occlusions and contacts appear where they occur in 3D; we add no separate camera conditioning, so camera motion that leaves the projected foreground unchanged is ambiguous.
We write the flow-matching path as $z^\tau=(1-\tau)\,z+\tau\,\epsilon$ with target velocity $v^\ast=\epsilon-z$, noise $\epsilon\sim\mathcal{N}(0,\mathbf{I})$ and diffusion time $\tau\in[0,1]$, on the shifted timestep grid and with the loss weights of the backbone.

\subsection{In-context control in the bidirectional teacher}
\label{sec:teacher}

\paragraph{Joint spatial packing}
We concatenate clean control latents and noisy target latents along the width axis,
\begin{equation}
h_i(\tau)=\operatorname{Concat}_W\big[\,u_i,\;z_i^\tau\,\big],
\label{eq:pack}
\end{equation}
where $i$ indexes latent time.
A packed frame has shape $48{\times}30{\times}108$, and with $2{\times}2$ spatial patches it contains 810 tokens, 405 for control and 405 for target.
Both halves share temporal positions, and the spatial positional encoding distinguishes their coordinates.
All visual attention operates on this joint sequence, so target features attend to the full control grid at every layer, while the output and the loss are restricted to the target half (\cref{fig:overview}a).
This in-context conditioning~\cite{incontextlora,ominicontrol,fulldit,vace} adds no conditioning branch, unlike adapters~\cite{controlnet} or new input channels~\cite{motionstream}, while all backbone weights are fine-tuned, and exposes the control at every layer, at the price of twice the visual tokens (\cref{sec:efficiency}).
In their settings, prior ablations found such token-level conditioning better than channel concatenation~\cite{scail,recammaster,camclonemaster} and feature addition~\cite{camclonemaster}.
Unlike OminiControl~\cite{ominicontrol}, which shares positions between condition and target tokens, and WorldWander~\cite{worldwander}, which concatenates them along the frame axis, packing along the width puts both halves of a time step into one latent frame, so that the block-causal mask over latent frames also makes the control causal, without separate positions or mask entries for control tokens.

\paragraph{Reference chunk}
The references form a clean chunk $R=[\,\varnothing,\,E(a_{\text{c}}),\,E(a_{\text{s}})\,]$ of three latent frames that precedes the video.
Each of its latents is packed with an all-zero control half, receives diffusion time zero and is excluded from the loss, and a missing reference leaves its slot at zero.
Three latent frames match the block size of the student, which keeps $R$ as its attention sink (\cref{sec:student}).
During training, the first frame and each reference are dropped independently with probability $0.1$, so the model renders from any subset of appearance inputs.

\paragraph{Teacher objective}
Control and reference tokens always have diffusion time zero, the anchor $z_0$ is restored after each sampling update, and classifier-free guidance~\cite{cfg} changes only the text condition.
Let $v_\phi$ denote the target-half prediction after joint processing.
The teacher minimizes
\begin{equation}
\mathcal{L}_{\mathrm{T}}=\mathbb{E}\!\left[\frac{\sum_{i\in\mathcal{M}} w(\tau)\,\big\|v_\phi\big(R,h(\tau),q\big)_i-v^\ast_i\big\|_2^2}{|\mathcal{M}|}\right],
\label{eq:teacher_loss}
\end{equation}
where $\mathcal{M}$ contains all target latents except $z_0$ when $I$ is given, and means are also taken over channels and space.

\subsection{Causal distillation and streaming inference}
\label{sec:student}

The teacher attends to future control frames, which violates R3, works on a fixed window, and samples with many denoising steps.
We convert it into a block-causal few-step student $\theta$ in two stages, following SolarWM~\cite{solarwm}: a teacher-forced flow-map initialization with AnyFlow~\cite{anyflow} (Stage~1), followed by distribution matching on the student's own rollouts~\cite{dmd,selfforcing,selfgradientforcing} (Stage~2).
Only the student is needed at inference, and it uses no classifier-free guidance.

\paragraph{Block-causal attention}
The student generates blocks of $b{=}3$ latent frames, \ie, $12$ video frames or $0.75$\,s.
Each block attends to the reference chunk, to its own packed frames and to up to $h{=}15$ preceding latent frames of clean history, and later blocks are masked out.
Because control and target of the same time step share one packed frame, the block-causal mask makes control and output causal together: block $k$ sees control only up to block $k$.
The reference chunk is visible to every block and is never evicted, so it acts as an attention sink~\cite{streamingllm} that carries the appearance inputs through sequences of any length.

\paragraph{Stage 1: causal initialization}
We first adapt the teacher weights to the block-causal student with the flow-map objective of AnyFlow~\cite{anyflow}.
For diffusion-time endpoints $\tau\ge r$, a learned flow map $\Phi_\theta(z^\tau;\tau,r)$ predicts the transition to $r$.
This stage uses ground-truth history (teacher forcing), the masks of \cref{eq:teacher_loss} and fused text guidance $3$.

\paragraph{Stage 2: distribution matching on generated histories}
We then roll out the student with its own committed outputs as context, as in Self Forcing~\cite{selfforcing}.
A frozen copy of the teacher supplies the reference score with text guidance of scale $3$, and a critic initialized from the teacher estimates the distribution of the noisy student samples; both receive the same control, references, anchor and text as the student.
At a sampled noise level, let $\hat z_{\text{real}}$ and $\hat z_{\text{fake}}$ be their clean-sample estimates.
Distribution matching~\cite{dmd,dmd2} supplies the target-masked direction
\begin{equation}
g=\frac{\hat z_{\text{fake}}-\hat z_{\text{real}}}{\max\big(\langle|\hat z_\theta-\hat z_{\text{real}}|\rangle_m,\,10^{-8}\big)},
\label{eq:dm_dir}
\end{equation}
where $\langle\cdot\rangle_m$ averages over the target elements $m$ of a sample, which exclude the anchor $z_0$; as in DMD~\cite{dmd}, we apply it through the surrogate $\tfrac12\|m\odot[\hat z_\theta-\operatorname{sg}(\hat z_\theta-g)]\|_2^2$, where $\operatorname{sg}$ stops gradients.
Following self-gradient forcing~\cite{selfgradientforcing}, rollouts are collected without graph and the student gradient comes from replaying a random denoising exit; the critic is updated at every iteration, the student at every fifth.

\paragraph{Few-step sampling and cache commitment}
The student uses four denoising levels ($[1000,750,500,250]$ mapped through the scheduler of the backbone) and re-noises each clean estimate with fresh noise to the next level; a zero-timestep pass then commits the keys and values of the clean chunk, so a chunk costs five transformer evaluations.
The cache holds the reference chunk, $15$ latent frames of history and the current block, and slides beyond the training window, so memory and cost per chunk stay constant; temporal positions are re-indexed within the window so that they never exceed the $24$ positions seen in training (supplementary material), and the anchor $z_0$ leaves with the history, whereas the reference chunk stays for the whole rollout.

\paragraph{Streaming interface}
The student factorizes the video causally across chunks $k$, $p_\theta(Z\mid U,I,A,q)=\prod_k p_\theta(Z_k\mid U_{\le k},\hat Z_{<k},I,A,q)$.
A chunk is $12$ frames, \ie, $0.75$\,s at $16$\,fps ($9$ frames for the first chunk), so the engine provides control one chunk ahead; if generation keeps up with real time and control is handed over without queuing, an input appears on screen after at most $0.75\,\text{s}+t_{\text{chunk}}$, where $t_{\text{chunk}}$ is the time to encode, generate and decode a chunk (R3).

\section{Data Construction}
\label{sec:data}

We construct aligned pairs $(C,Y)$ with a first frame and a character and a scene reference from three sources: $6{,}043$ engine clips, $5{,}344$ gameplay clips and $13{,}568$ external clips.

\paragraph{Engine renders: one game state, two rendering passes}
We script engine scenes of approach, attack, evasion, contact and separation with a controllable camera and drive low-poly proxies and detailed assets with the same state sequence.
One pass renders the proxies in uniform white on black to obtain $C$, a second the detailed assets with materials and lighting to obtain $Y$, with identical projection and frame times; the references are rendered from the same assets.
The training clips cover $172$ characters, $40$ weapons, three environments and $686$ choreography groups; two further characters are held out for testing (\cref{sec:setup}).

\paragraph{Gameplay: footage to reconstructed proxies}
We select gameplay clips of one-on-one melee combat from five commercial action games, \emph{Sekiro: Shadows Die Twice}, \emph{Where Winds Meet}, \emph{Wuchang: Fallen Feathers}, \emph{Sifu} and \emph{Rise of the Ronin}, and remove cuts, overlays and frames unsuitable for motion recovery.
We track every fighter, recover 3D articulation, root motion and camera, retarget the motion to proxy bodies and render $C$, paired with the footage as $Y$; a character sheet and a background plate are generated from up to three frames of the footage with an image-editing model ($47$ failures are removed).
The footage is used for research only and is not redistributed. %

\paragraph{External pairs}
We further use videos from seven external datasets: ABot~\cite{abot}, GTA-Human~II~\cite{gtahuman}, JTA~\cite{jta}, MultiCamVideo~\cite{recammaster}, MIND~\cite{mind}, MiraData~\cite{miradata} and OmniWorld~\cite{omniworld}.
For JTA and GTA-Human~II we drive volumetric proxies with the provided 3D joints or SMPL-X parameters; for the others we lift tracked 2D poses to 3D~\cite{motionagformer}, fit body proxies and render them with the annotated cameras.
Their references are crops of the paired video.

\section{Experiments}
\label{sec:exp}

\subsection{Setup}
\label{sec:setup}

\paragraph{Test set}
We test on held-out clips that were used neither for training nor for selecting checkpoints: $136$ engine renders of two characters that do not appear in training and fight each other in every clip, and $108$ clips of four external datasets (\cref{sec:data}).
Checkpoints were selected on other held-out clips, which show the same two characters and are excluded from all results; $100$ further unused clips pair one of the two with $18$ training characters (\emph{mixed pairs}), and $48$ clips rendered after all models were fixed show eight characters used neither in training nor in selection (\emph{new characters}).
The choreographies of the engine renders occur in training with other characters ($6$ test clips share a trajectory with a training clip; supplementary material).
Comparisons with other methods use $174$ clips, $66$ engine renders drawn at random and all external clips; ablations and one-minute rollouts use the engine renders.
Each clip provides the proxy video, the ground-truth video, its first frame $I$, a character and a scene reference and two captions: the \emph{clip caption} describes its motion, and the causal students receive the \emph{static caption} of the scene and the characters without motion; we evaluate the first $5$\,s ($81$ frames).

\paragraph{Implementation details}
We use the teacher after $6{,}000$ steps, Stage~1 after $3{,}000$ and Stage~2 after $2{,}000$ iterations, selected by PSNR on held-out validation clips (supplementary material).
At test time, the teacher samples with $50$ Euler steps and text guidance $3$, Stage~1 with four flow-map steps, and Stage~2 with four steps plus the commit pass.
Timings are on one A100 in bfloat16.

\paragraph{Metrics}
We compare generated and ground-truth frames with MSE, PSNR, SSIM and LPIPS~\cite{lpips}, excluding the given first frame, after resampling every output to $16$\,fps and $480{\times}864$.
Pixel metrics reward blur and miss identity errors, so a multimodal language model also rates \emph{motion} following, \emph{appearance} consistency and visual \emph{quality} from $1$ to $5$ from four frames of control and output and the appearance inputs, without method names and in two independent requests, and names failures such as a missing weapon.

\paragraph{Baselines}
None of the compared methods takes our proxy render as its native control, so each receives the control it was trained with: depth from Video Depth Anything~\cite{videodepthanything} and G-buffers with this depth, its normals and a constant white albedo, computed from the proxy video because engine depth was not exported, and DWPose~\cite{dwpose} skeletons estimated on the ground-truth frames, because DWPose finds both fighters in fewer than $7\%$ of proxy frames.
We compare with two neural game renderers, Marionette~\cite{marionette}, which turns a pose-control video and a first frame into combat-game footage, and the offline G-buffer renderer Generative World Renderer~\cite{genworldrenderer}, as well as with Wan2.2-Fun-5B-Control~\cite{videoxfun}, which shares our backbone and accepts a skeleton, a depth map or our proxy video.
Every baseline that accepts an image receives the same first frame $I$, and none accepts character or scene references, so we also report our model with the first frame only; all models have at most $5$B parameters.
Magpie~\cite{magpie} and Wizard~\cite{wizard} have released no weights, and SolarRenderer~\cite{solarrenderer} is released for a far larger backbone.
The existing renderers are used as released, without training on our data, so this comparison measures systems; \cref{sec:representation} examines the control.

\subsection{Comparison and training stages}
\label{sec:main_results}

\newlength{\qlw}\setlength{\qlw}{0.445\textwidth}
\newcommand{\qlstrip}[2]{\raisebox{-0.5\height}{\includegraphics[width=\qlw]{fig/img/q_#1_#2.jpg}}}
\newcommand{\qltime}[1]{\makebox[\dimexpr\qlw*640/2578\relax]{\scriptsize #1}}
\newcommand{\qlgap}{\hspace{\dimexpr\qlw*6/2578\relax}}
\newcommand{\qlheads}{\qltime{$0.5$\,s}\qlgap\qltime{$1.5$\,s}\qlgap\qltime{$3.0$\,s}\qlgap\qltime{$4.5$\,s}}
\newcommand{\qlrow}[3][]{\rowlab[#1]{#2}\hspace{3pt}\qlstrip{#3}{A}\\[1.5pt]}
\newlength{\mcw}\setlength{\mcw}{0.0707\textwidth}
\newcommand{\mct}[2]{\includegraphics[width=\mcw]{fig/img/mc_#1_#2.jpg}}
\newcommand{\mcrow}[1]{\mct{ctl}{#1} & \mct{funskel}{#1} & \mct{stage2}{#1} & \mct{tskel}{#1} & \mct{tproxy}{#1} & \mct{gt}{#1}}
\newcommand{\mcsub}[1]{{\scriptsize\color{black!55}#1}}
\begin{figure*}[t]
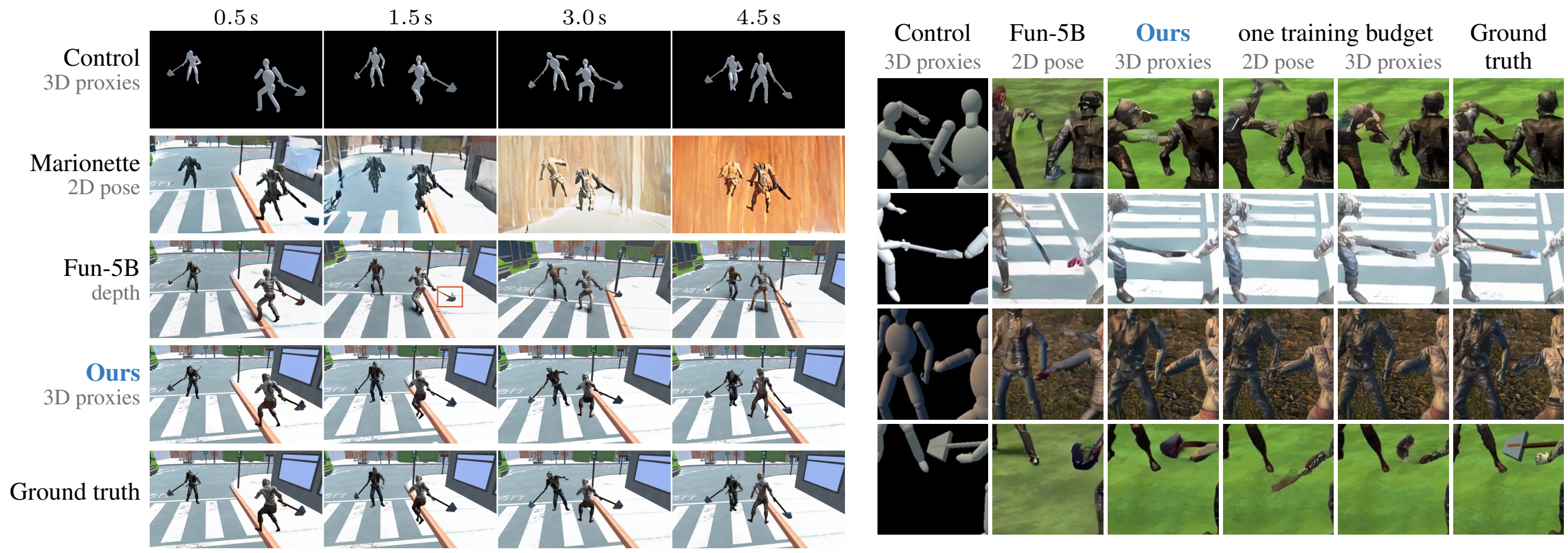

\centering
\begin{minipage}[c]{0.545\textwidth}
\raggedright
\rowlab{\mbox{}}\hspace{3pt}\qlheads\\[1pt]
\qlrow[3D proxies]{Control}{clay}
\qlrow[2D pose]{Marionette}{marionette}
\qlrow[depth]{Fun-5B}{fun5b_depth}
\qlrow[3D proxies]{\ours}{ours_stage2}
\rowlab{Ground truth}\hspace{3pt}\qlstrip{gt}{A}
\end{minipage}\hfill
\begin{minipage}[c]{0.44\textwidth}
\raggedleft
\setlength{\tabcolsep}{0.75pt}\renewcommand{\arraystretch}{0}%
\begin{tabular}{@{}cccccc@{}}
{\footnotesize Control} & {\footnotesize Fun-5B} & {\footnotesize\ours} & \multicolumn{2}{c}{\footnotesize one training budget} & {\footnotesize Ground}\\[3.5pt]
\mcsub{3D proxies} & \mcsub{2D pose} & \mcsub{3D proxies} & \mcsub{2D pose} & \mcsub{3D proxies} & {\footnotesize truth}\\[3pt]
\mcrow{0}\\[1.5pt]
\mcrow{1}\\[1.5pt]
\mcrow{2}\\[1.5pt]
\mcrow{3}
\end{tabular}
\end{minipage}
\caption{\textbf{Qualitative comparison.} Left: an engine test clip; the box marks a weapon that Fun-5B-Control changes. Right: windows at four contact events chosen from the control alone, with the two teachers of \cref{sec:representation}. With skeletons the weapon is lost or misplaced; with proxies it follows the control but is darker and less detailed than the ground truth, and barely visible in the first event.}
\label{fig:qualitative}
\end{figure*}

\begin{table*}[t]
\centering
\small
\caption{\textbf{Comparison} on $66$ engine renders and $108$ external clips. Motion, App.: ratings from $1$ to $5$. Steps: evaluations per step $\times$ steps. Time: seconds for $5$\,s of video on one A100 ($^\ddagger$sliding window; $16.6$ with the KV cache of Stage~2). $^\dagger$On $80$ clips.}
\label{tab:main}
\setlength{\tabcolsep}{3.2pt}
\resizebox{\linewidth}{!}{%
\begin{tabular}{@{}ll*{11}{c}@{}}
\toprule
& & & & \multicolumn{4}{c}{Engine renders} & \multicolumn{4}{c}{External clips} & \\
\cmidrule(lr){5-8}\cmidrule(lr){9-12}
Control & Method & Causal & Steps & PSNR$\uparrow$ & LPIPS$\downarrow$ & Motion$\uparrow$ & App.$\uparrow$ & PSNR$\uparrow$ & LPIPS$\downarrow$ & Motion$\uparrow$ & App.$\uparrow$ & Time \\
\midrule
\multirow{2}{*}{2D pose} & Marionette~\cite{marionette} & \xmark & $2{\times}40$ & 14.49 & 0.544 & 2.87 & 2.77 & 13.54$^\dagger$ & 0.587$^\dagger$ & 2.58$^\dagger$ & 2.81$^\dagger$ & 73.5 \\
& Fun-5B-Control~\cite{videoxfun} & \xmark & $2{\times}50$ & 16.62 & 0.402 & 3.55 & 4.00 & 15.46 & 0.490 & 3.68 & 4.39 & 59.3 \\
\midrule
Depth & Fun-5B-Control~\cite{videoxfun} & \xmark & $2{\times}50$ & 16.46 & 0.407 & 4.11 & 4.05 & 14.95 & 0.513 & 3.83 & 4.24 & 59.5 \\
\midrule
G-buffer & World Renderer~\cite{genworldrenderer} & \xmark & $2{\times}50$ & 9.09 & 0.851 & 3.81 & 1.14 & 9.87 & 0.856 & 3.52 & 1.00 & 212.2 \\
\midrule
\multirow{5}{*}{3D proxies} & Fun-5B-Control~\cite{videoxfun} & \xmark & $2{\times}50$ & 16.46 & 0.407 & 4.28 & 3.95 & 15.04 & 0.516 & \textbf{3.92} & 4.18 & 59.0 \\
& Ours, teacher & \xmark & $2{\times}50$ & \textbf{21.31} & \textbf{0.146} & \textbf{4.52} & \textbf{4.69} & \textbf{15.61} & \textbf{0.464} & 3.85 & \textbf{4.69} & 195.0 \\
& Ours, Stage~1 & \cmark & $1{\times}4$ & 19.37 & 0.218 & 4.45 & 4.58 & 14.14 & 0.589 & 3.70 & 4.37 & 34.3$^\ddagger$ \\
& Ours, Stage~2 & \cmark & $1{\times}4{+}1$ & 20.31 & 0.184 & 4.51 & 4.67 & 12.79 & 0.611 & 3.66 & 4.35 & 16.6 \\
& \quad first frame only & \cmark & $1{\times}4{+}1$ & 20.04 & 0.201 & 4.44 & 4.64 & 12.05 & 0.636 & 3.57 & 4.02 & 16.6 \\
\bottomrule
\end{tabular}}
\end{table*}

On engine renders (\cref{tab:main}, \cref{fig:qualitative}), all our models clearly outperform the other renderers: Stage~2 reaches $20.31$\,dB PSNR and LPIPS $0.184$, against $16.62$\,dB and $0.402$ for the best baseline, is ahead of all five baselines in both measures on each of the $66$ clips, and remains so with the first frame as its only appearance input, like the baselines ($20.04$\,dB).
On the mixed pairs, with other partners, weapons and trajectories, it reaches $19.81$\,dB and a motion rating of $4.14$; on the new characters, $19.49$\,dB and $4.23$ (supplementary material).
Marionette, trained on footage of one game, loses the scene and the weapons, which our reconstruction of its control does not contain.
World Renderer, given constant material maps and no image, renders untextured figures (appearance $1.14$), which reflects our inputs, not G-buffers as a control.
The external clips show no combat, and their proxies are reconstructed from video and cover only the human bodies, so the control leaves the motion of the scene open. There the teacher still gives the best PSNR, LPIPS and appearance rating, so neither the reconstructed proxies nor the domain alone explain the loss of the students: it arises in both distillation stages ($1.5$ and $1.3$\,dB), is largest in the first chunk and does not grow, leaving Stage~2 $2.7$\,dB behind Fun-5B-Control with skeletons.

\subsection{Control following and contact events}
\label{sec:representation}

\emph{Control swap.}
When Stage~2 receives the control of another clip, its output follows that control and not the original motion: the PCK of body joints (DWPose on all $80$ frames) is $0.88$ against the poses of the clip that supplied the control and $0.38$ against those of the original clip, and an empty control lowers the motion rating from $4.49$ to $2.09$ (\cref{tab:ablation}b).

\emph{Proxies or skeletons.}
We train two teachers that differ only in the control: both start from the same initialization and see the same $1{,}000$ engine clips for $4{,}000$ steps, one with proxies and one with skeletons estimated from the ground-truth frames.
On $33$ test clips their PSNR does not differ ($20.59$ vs.\ $20.47$; difference $+0.11$, $95\%$ interval $-0.10$ to $0.32$), but the skeleton model has the lower motion rating ($4.39$ vs.\ $4.64$; $+0.24$, $0.08$ to $0.41$), the rater names a missing weapon in $18\%$ of its ratings against $2\%$, and at the contact events below its strike correlation is $0.30$ against $0.58$ ($+0.28$, $0.18$ to $0.41$) and exceeds $0.5$ in $30\%$ of the events against $73\%$; the gaps are similar after $1{,}000$ and $2{,}000$ steps (supplementary material).
A skeleton carries no props, and training does not recover them (\cref{fig:qualitative}, right).
A released model shows the same: Fun-5B-Control, never trained with proxies, has a higher motion rating with proxies than with skeletons ($4.28$ vs.\ $3.55$) and loses the weapon in $27\%$ of the skeleton ratings against $2\%$.
For our full model, a flat silhouette of the proxies stays close to the shaded render in the averages (PSNR $19.97$ vs.\ $20.10$, motion $4.39$ vs.\ $4.49$), with more wrong poses ($47\%$ vs.\ $37\%$ of the ratings) and, in the videos we inspected, weapons that lose their shape during strikes.

\emph{Contact events.}
From the control alone, we locate the frames at which the silhouettes of the two characters merge, \ie, a weapon or limb meets the opponent in the image ($72$ events), and evaluate a window of $128$ pixels over the seven frames around each.
These are contacts in the image: by the engine's record of $54$ further clips, $65\%$ of such events fall into an attack or its reaction and the others into an approach, and the two come within $10$\,cm in 3D at $36\%$ (of all windows: $10\%$; supplementary material).
In all of the events, the window of Stage~2 matches the ground truth best without a time offset, a measure of local alignment that also depends on appearance (Fun-5B-Control: $53$ to $56$ events), and the two identities are correct in $99\%$ of these frames, against $82$ to $86\%$.
The change at the pixels of the striking part correlates with that of the ground truth by $0.59$ on average, and the strike \emph{appears}, with a correlation above $0.5$, in $68\%$ of the $69$ events with a strike mask, against $4$ to $14\%$ for the five baselines (at most $0.19$ on average), at weapon strikes ($71\%$ against at most $12\%$) as at contacts of limbs ($62\%$ against at most $38\%$).
A flat silhouette nearly reproduces these values, which do not show whether the striking part is rendered in full, so we also ask the rater, blindly, whether it is visible and reaches the opponent.
Where the part is a weapon ($48$ events), it is missing (absent or elsewhere) in $17\%$ of the answers for Stage~2, against $38$ to $58\%$ for Fun-5B-Control, and fully visible in $36\%$, against $4$ to $15\%$ (ground truth: never missing, fully visible in $91\%$): the baselines often lose the weapon; ours keeps it but renders it fully in only a third of the events.
Ten volunteers who are not authors, asked the same question blind to the method, agree at weapon strikes (\cref{tab:human}); at the $24$ contacts of limbs, where the rater is unreliable ($48\%$ for the ground truth), they see the limb fully in $84\%$ of their answers for Stage~2, against $63$ and $54\%$.

\begin{table}[t]
\centering
\small
\caption{\textbf{Blind human check} at the contact events: answers (\%) of ten volunteers on whether the striking part is visible and reaches the opponent (yes, partly, no).}
\label{tab:human}
\setlength{\tabcolsep}{3.2pt}
\resizebox{\linewidth}{!}{%
\begin{tabular}{@{}lcccccc@{}}
\toprule
& \multicolumn{3}{c}{Weapon ($48$ events)} & \multicolumn{3}{c}{Limb ($24$ events)} \\
\cmidrule(lr){2-4}\cmidrule(lr){5-7}
Method & yes & partly & no & yes & partly & no \\
\midrule
Ground truth & 95 & 5 & 0 & 95 & 1 & 4 \\
Fun-5B-Control, skeleton & 11 & 24 & 65 & 54 & 33 & 13 \\
Fun-5B-Control, proxies & 12 & 37 & 51 & 63 & 29 & 8 \\
Ours, Stage~2 & \textbf{31} & 53 & \textbf{16} & \textbf{84} & 14 & \textbf{2} \\
\bottomrule
\end{tabular}}
\end{table}

\subsection{Long sequences}
\label{sec:long}

Lacking one-minute recordings, we play each of $19$ test clips forward and backward for $60.6$\,s and generate it with Stage~2 in one rollout without resetting the cache.
From the first to the last $10$\,s, the appearance rating drops from $4.45$ to $4.32$, the DINOv2 similarity to the first frame from $0.888$ to $0.861$ and PSNR from $18.39$ to $15.33$.
Without references, the late appearance rating falls to $2.61$ and PSNR to $11.93$: the first frame leaves the cache after $4.3$\,s, and the references are then the only anchor.
With the clip caption, Stage~1 with the same cache drifts more than Stage~2 (late PSNR $12.90$ vs.\ $14.88$, appearance $3.47$ vs.\ $4.05$).
With one denoising step (\cref{sec:efficiency}), the late PSNR ($15.59$) and appearance ($4.29$) hold, and the late quality rating is $0.3$ lower.

\subsection{Efficiency}
\label{sec:efficiency}

On one A100, a chunk of $12$ frames takes $t_{\text{chunk}}=2.67$\,s with control encoding, decoding and the copy to the host, of which $1.78$\,s are spent in the transformer and $0.72$\,s in decoding.
Stage~2 thus renders $4.5$ frames per second, $3.6{\times}$ below real time, with a constant peak memory of $38$\,GB and the first frame after $1.8$\,s.
To measure the \emph{switch latency}, we swap in the control of another clip at a chunk boundary: all frames before the switch are bit-identical to those of the unchanged rollout, and the change appears after $2.7$\,s, in the first frame of the switched chunk.
With one denoising step and control encoding, transformer and decoding as a pipeline on three A100s, the renderer reaches $17.3$ frames per second, above the $16$ of real time, and the strike appears in $67\%$ of the events (four steps: $68\%$). Queues between stages raise the switch latency to $3.7$\,s ($1.6$\,s in the stages).

\subsection{Inference-time ablations}
\label{sec:ablation}

\begin{table}[t]
\centering
\small
\caption{\textbf{Inference-time ablations} of Stage~2 on the $136$ engine test renders with the clip caption, without retraining (defaults \colorbox{linkblue!12}{highlighted}).}
\label{tab:ablation}
\setlength{\tabcolsep}{3pt}
\resizebox{\linewidth}{!}{%
\begin{tabular}{@{}llcccc@{}}
\toprule
& Variant & PSNR$\uparrow$ & LPIPS$\downarrow$ & Motion$\uparrow$ & App.$\uparrow$ \\
\midrule
\multirow{4}{*}{(a)} & no appearance input & 12.66 & 0.603 & 4.15 & 2.68 \\
& references only & 19.33 & 0.196 & 4.38 & 4.29 \\
& first frame only & 19.87 & 0.198 & 4.43 & \textbf{4.67} \\
& \cellcolor{linkblue!12}first frame + references & \cellcolor{linkblue!12}\textbf{20.10} & \cellcolor{linkblue!12}\textbf{0.186} & \cellcolor{linkblue!12}\textbf{4.49} & \cellcolor{linkblue!12}4.60 \\
\midrule
\multirow{5}{*}{(b)} & empty control & 16.16 & 0.410 & 2.09 & 3.99 \\
& control of another clip & 17.18 & 0.335 & 3.12 & 3.83 \\
& skeleton (ground-truth frames) & 19.14 & 0.229 & 3.02 & 3.41 \\
& flat silhouette & 19.97 & 0.190 & 4.39 & 4.60 \\
& \cellcolor{linkblue!12}3D proxies & \cellcolor{linkblue!12}\textbf{20.10} & \cellcolor{linkblue!12}\textbf{0.186} & \cellcolor{linkblue!12}\textbf{4.49} & \cellcolor{linkblue!12}4.60 \\
\bottomrule
\end{tabular}}
\end{table}

\cref{tab:ablation} varies the inputs of Stage~2 (block~(b): \cref{sec:representation}); its rows use the clip caption, which a streaming renderer would not have, but the static caption gives the same results (PSNR $20.31$ vs.\ $20.27$).
Without any appearance input (block~(a)) the appearance rating falls to $2.68$; references alone raise it to $4.29$, the first frame alone gives most of the fidelity (PSNR $19.87$), and both together give the best PSNR and LPIPS.
With two or one denoising steps, PSNR stays at $20.23$ and $20.05$ ($6.2$ and $7.5$ frames per second).

\section{Conclusion}
\label{sec:conclusion}

We presented \method, a streaming neural renderer behind a game engine: a teacher packs control and target into the same latent frames, and a block-causal few-step student reads the control chunk by chunk and keeps the references as an attention sink.
On engine renders of two unseen characters, the student stays within $1.0$\,dB of its teacher, follows a control switch from the next chunk and stays aligned with the ground truth at contact events; it renders $4.5$ frames per second on one A100 and $17$ with one step on three, and sustains one-minute rollouts with its references as anchor.
Released renderers, not trained on our data, stay $3.7$\,dB below it; in a blind human check, one of them loses a striking weapon in $51$ and $65\%$ of the answers with our proxies and skeletons, ours in $16\%$.

\paragraph{Limitations}
Our engine evaluation uses recorded control of close combat between characters of one asset family, and the one-minute rollouts replay short clips; closed-loop play with a running engine, other asset families and other genres remain untested.
Real time currently takes three GPUs and comes with a switch latency of $3.7$\,s.
Our release will carry a usage policy against depicting real people.

\section*{LLM Usage Statement}
The authors used large language models to draft and edit text, to search and summarize related work, to write code for data processing, evaluation and figures, to generate the reference images of the gameplay clips, and as an automatic rater in the evaluation (\cref{sec:setup}).
The authors designed the method and the experiments, checked all statements, citations and numbers, and take full responsibility for the content.

{
    \small
    \bibliographystyle{ieeenat_fullname}
    \bibliography{refs}
}

\clearpage
\setcounter{section}{0}
\setcounter{figure}{0}
\setcounter{table}{0}
\setcounter{equation}{0}
\setcounter{algorithm}{0}
\renewcommand{\thesection}{S\arabic{section}}
\renewcommand{\thefigure}{S\arabic{figure}}
\renewcommand{\thetable}{S\arabic{table}}
\renewcommand{\theequation}{S\arabic{equation}}
\renewcommand{\thealgorithm}{S\arabic{algorithm}}
\makeatletter
\renewcommand{\theHsection}{S\arabic{section}}
\renewcommand{\theHfigure}{S\arabic{figure}}
\renewcommand{\theHtable}{S\arabic{table}}
\renewcommand{\theHequation}{S\arabic{equation}}
\def\theHalgorithm{S\arabic{algorithm}}
\makeatother
\maketitlesupplementary

\section{Overview}
\label{supp:overview}
This supplementary material provides
implementation details of the teacher, both student stages and inference (\cref{supp:impl}),
data details (\cref{supp:data}),
evaluation protocols, including the multimodal rating and the one-minute sequences (\cref{supp:eval}),
baseline protocols (\cref{supp:baselines})%
,
additional results (\cref{supp:results}),
and a description of the supplementary video (\cref{supp:videos}).

\section{Implementation details}
\label{supp:impl}

\paragraph{Teacher}
The teacher is initialized from the released Wan2.2-TI2V-5B transformer~\cite{wan2025} and trained with the bidirectional flow-matching stage of SolarWM~\cite{solarwm}: one diffusion time per window on the shifted grid, with the loss weights of the backbone and the masks of \cref{eq:teacher_loss}.
Each GPU processes one window, and $81$-frame and $45$-frame windows are mixed across iterations.
The first frame and each reference are dropped with probability $0.1$.
We use AdamW with learning rate $10^{-5}$, $\beta=(0.9,0.95)$, weight decay $0.01$, $200$ warm-up steps and cosine decay to $0.1\times$, gradient clipping at $1$, and an EMA with decay $0.999$, whose weights we use for evaluation and for the students.
We evaluate checkpoints at $3{,}000$ to $6{,}000$ steps with $50$-step sampling and select $6{,}000$ steps by PSNR on $64$ held-out engine clips and $64$ held-out external clips; all of them are excluded from the evaluation (\cref{supp:eval}).

\paragraph{Stage 1}
The student is initialized from the EMA teacher, with the AnyFlow interval embedding copied from the time embedding.
Training uses teacher forcing over the full window with blocks of $3$ latent frames, a history of $15$ latent frames and the reference chunk as sink, the AnyFlow v1.5 objective of SolarWM with fused text guidance $3$, learning rate $3\cdot10^{-5}$ and EMA $0.999$.
We evaluate checkpoints at every $1{,}000$ steps with four flow-map steps and select $3{,}000$ steps by PSNR on the same validation clips as the teacher.

\paragraph{Stage 2}
The student starts from the EMA of Stage~1 without the interval embedding; the frozen teacher and the critic both start from the EMA teacher.
Each iteration updates the critic with flow matching on a detached student rollout, and every fifth iteration also updates the student.
Student rollouts cover the $21$-latent window with the KV cache, four levels from $[1000,750,500,250]$ and a random exit level; the gradient is obtained by replaying the exit prediction under teacher forcing on the committed chunks~\cite{selfgradientforcing}.
Score timesteps are sampled on the shifted grid in $[20,980]$, the real score uses text guidance $3$, and the direction of \cref{eq:dm_dir} is normalized.
We use Adam with $\beta=(0,0.999)$, learning rates $2\cdot10^{-6}$ for the student and $4\cdot10^{-7}$ for the critic, gradient clipping at $10$ and an EMA of $0.99$ that starts after $39$ student updates.
We train for $2{,}500$ iterations and evaluate candidates on eight of the selection clips, four of them additionally as $60.6$-s forward--backward rollouts; the score $0.25\,\mathrm{PSNR}_{\text{short}}+0.25\,\mathrm{PSNR}_{\text{middle}}+0.5\,\mathrm{PSNR}_{\text{late}}-0.5\max(0,\mathrm{PSNR}_{\text{early}}-\mathrm{PSNR}_{\text{late}})$ selects $2{,}000$ iterations.

\paragraph{Inference}
Stage~1 samples each block with four flow-map steps, and Stage~2 uses the KV cache of \cref{alg:stream}. The Stage~1 outputs of \cref{tab:main} were sampled over a cache-free sliding window that recomputes the history in each of the four steps ($34.3$\,s per clip, the time in \cref{tab:main}); its one-minute rollouts use the KV cache of Stage~2, with which $5$\,s of video take $16.6$\,s.
Once the cache is full, the rotary temporal positions are re-indexed within the window: the reference chunk keeps positions $0$ to $2$, positions $3$ to $5$ stay empty, and the $15$ history frames and the current block take positions $6$ to $23$, the end of the $24$-position horizon of training; we use no other positional extrapolation.
The cache stores keys and values of $21$ packed latent frames, \ie, $21\cdot810=17{,}010$ visual tokens per layer; with $30$ layers and hidden size $3072$, this is ${\approx}6.3$\,GB in bf16, independent of the sequence length, of a measured peak of $38.1$\,GB (\cref{tab:profile}).

\begin{algorithm}[h]
\caption{Streaming inference of \method}
\label{alg:stream}
\small
\begin{algorithmic}[1]
\Require control chunks $C_0,C_1,\dots$ from the engine; optional first frame $I$ and references $A$; text $q$; levels $\tau_1{>}\tau_2{>}\tau_3{>}\tau_4$
\State $R\gets[\varnothing,E(a_{\text{c}}),E(a_{\text{s}})]$;\ \ encode $q$ once
\State $\mathcal{K}\gets\mathrm{KV}_\theta(R,0;q)$ \Comment{attention sink}
\For{$k=0,1,2,\dots$}
  \State $U_k\gets E(C_k)$ \Comment{control of chunk $k$ only}
  \State $Z_k\gets\epsilon\sim\mathcal{N}(0,\mathbf{I})$ \Comment{latent $0\gets E(I)$ if given}
  \For{$j=1,\dots,4$}
    \State $\hat Z_k\gets\hat z_\theta\big([U_k\,|\,Z_k],\tau_j;\mathcal{K},q\big)$ \Comment{clean estimate}
    \State \textbf{if} $j<4$: re-noise $\hat Z_k$ to level $\tau_{j+1}$
  \EndFor
  \State $\mathcal{K}\gets\mathcal{K}\cup\mathrm{KV}_\theta\big([U_k\,|\,\hat Z_k],0;\mathcal{K},q\big)$ \Comment{commit}
  \State keep $R$ and the last $h{=}15$ latent frames in $\mathcal{K}$
  \State decode $\hat Z_k$ with the causal decoder and emit its frames
\EndFor
\end{algorithmic}
\end{algorithm}

\section{Data details}
\label{supp:data}
\cref{tab:data} lists the training clips by source.

\begin{table}[h]
\centering
\small
\caption{\textbf{Training data by source.}}
\label{tab:data}
\setlength{\tabcolsep}{4pt}
\resizebox{\linewidth}{!}{%
\begin{tabular}{@{}lccc@{}}
\toprule
Source & Clips & Control & References \\
\midrule
Engine renders & 6{,}043 & exact & rendered \\
Gameplay (5 games) & 5{,}344 & reconstructed & generated \\
External (7 datasets) & 13{,}568 & annotated or reconstructed & cropped \\
\bottomrule
\end{tabular}}
\end{table}

Frames are sampled at $t=t_{\text{start}}+n/16$\,s by nearest presentation time, resized to cover $480{\times}864$ and center-cropped; sources slightly below $16$\,fps are taken frame by frame.
Fifteen-second clips give windows starting at $0$, $4.9$ and $9.9$\,s, and clips shorter than $5$\,s give one $45$-frame window.
Captions are written in English by DeepSeek-V4.1-Flash from the control and the references, without the target video, and describe only visible content; the first window of a clip uses its clip caption and later windows a static scene caption. An example: \emph{``Both figures raise their weapons into wide braced stances, then close distance. The sword-bearer lowers to a diagonal guard while the polearm-bearer swings down and across.''}
Engine clips come with a character reference and a scene reference rendered from the same assets.
Gameplay clips have no clean assets: an image-editing model receives one to three frames of the footage and produces a character sheet (the player from behind and the opponent from the front, full body with weapons, on a plain background) and a background plate without characters, effects or interface elements; outputs that cannot be decoded, are smaller than $256$ pixels or miss either image are retried, and the $47$ clips that still fail are removed.
For the external datasets, JTA proxies are driven by the provided 3D joints and GTA-Human~II proxies by the provided SMPL-X parameters; for ABot, MultiCamVideo, MIND, MiraData and OmniWorld, YOLO11-pose detections are tracked, lifted to 3D with MotionAGFormer~\cite{motionagformer}, fitted with fixed bone lengths and a perspective root, and rendered as volumetric proxies with a shared depth buffer and the annotated camera.
Their references are rectangular crops of the paired video: the characters from five frames and the scene from the largest region without characters.
Of the $13{,}858$ external clips, $290$ (two percent, chosen by a hash of the clip name) are held out and form the external part of the held-out pool (\cref{supp:eval}); two of the training clips come from an eighth dataset, FORM-HOI\footnote{\url{https://huggingface.co/datasets/nvidia/form-hoi}}.
The $6{,}043$ engine training clips cover $172$ characters ($178$ asset records before merging known aliases), $40$ weapons, three environments with $31$ sub-scenes and $686$ of the $700$ choreography groups.
The two test characters appear in $1{,}242$ engine clips: $200$ form the test set and the other $1{,}042$ are not used for training ($381$ in which they fight each other and $661$ in which one of them fights a training character).

\section{Evaluation details}
\label{supp:eval}

\paragraph{Test set}
The held-out pool contains $200$ engine renders and $290$ external clips, from which we sampled $140$ external clips, stratified by dataset, with a fixed seed. Checkpoints were selected on the first $64$ engine and the first $64$ external clips of this pool in a fixed order; we exclude them, which removes $64$ engine renders and $32$ of the sampled external clips (all ABot, JTA and GTA-Human~II clips and two MultiCamVideo clips). The remaining test set has $136$ engine renders and $108$ external clips from MultiCamVideo ($70$), OmniWorld ($28$), MIND ($9$) and MiraData ($1$).
The baselines were run on $100$ engine renders, the first $100$ of a random permutation (seed $0$) of the sorted clip identifiers, and on the $140$ external clips; comparisons use the $66$ engine renders and $108$ external clips among them that were not used for selection.
The external clips show one or a few characters walking, posing or riding through a scene under a moving camera, not combat; their proxies are reconstructed from the video and cover only the human bodies, so the control does not determine how the scene moves.
They test generalization beyond our target task.

\paragraph{Independent re-test}
To check that the test set is not favorable by chance, we drew $100$ further engine clips at random (fixed seed) from the $381$ unused clips in which the two test characters fight each other; they were never used for training, validation or selection.
Stage~2 reaches PSNR $20.32$, LPIPS $0.182$, motion $4.58$ and appearance $4.62$ on them, against $20.10$, $0.186$, $4.49$ and $4.60$ on the $136$ test clips.

\paragraph{Further character combinations}
\label{supp:mixed}
All test clips show the same pair of characters, and so do the clips on which checkpoints were selected: the test clips are independent of the selection, the two identities are not.
To test other combinations, we use the $661$ unused clips in which one of the two test characters fights a training character; none of them was used for training, validation or selection, and none of their motion trajectories occurs in a training clip.
We drew at most eight clips for each of the $18$ partners with a fixed seed, $100$ clips in total; in $92$ of them the test character carries one of three weapons that it does not carry in the test set, and the partners carry $14$ different weapons.
Stage~2 reaches PSNR $19.81$ $[19.49, 20.13]$, LPIPS $0.218$, motion $4.14$ $[4.05, 4.24]$ and appearance $4.46$ on them with the static caption, against $20.31$, $0.184$, $4.51$ and $4.67$ on the $66$ comparison clips: the fidelity is close, and the ratings are lower with new partners and weapons.
These clips widen the set of pairs, weapons and trajectories but add no unseen identity.

\paragraph{New characters}
\label{supp:newid}
After all models were fixed, we rendered $48$ clips with eight further characters from the same asset family that occur in no training, validation or selection clip (asset files, aliases and meshes were compared with all characters used before). Every clip shows two of them; each character appears in $12$ clips with seven partners, and the three environments have $16$ clips each. Pairs, weapons and choreographies were drawn by the original generator with new seeds; configurations that it rejected were redrawn automatically, and no clip was removed after rendering. The three clips of \cref{fig:teaser} are from this set, one per environment. The rendering project was restored from a backup, so six clips of the two test characters were rendered with it as a calibration.
Stage~2 with the static caption reaches PSNR $19.49$ $[18.85, 20.16]$, LPIPS $0.243$, motion $4.23$ $[4.07, 4.39]$ and appearance $4.43$ on the new characters, and $20.54$, $0.198$, $4.58$ and $4.67$ on the calibration clips, which agree with the test set ($20.31$, $0.184$, $4.51$, $4.67$). At the $64$ contact events of the new clips its window is aligned in $62$, with a strike correlation of $0.62$ $[0.52, 0.72]$; the strike appears (correlation above $0.5$, \cref{supp:contact}) in $70\%$ of the $50$ events with a strike mask, as for the test characters ($68\%$ of $69$); a mask exists for fewer events here ($78\%$ against $96\%$), so the two shares describe the measurable events of each set. New characters thus cost about $0.8$\,dB and $0.3$ rating points against the two test characters, and the rater names an identity swap in $6\%$ and a missing weapon in $7\%$ of the ratings (test set: $0\%$ and $1\%$).

\paragraph{Overlap between test and training data}
No test clip is used for training.
In every engine test clip the same two characters fight each other; they are excluded from training together with their other $1{,}042$ clips, and no training character shares an asset, a known alias or a display name with them.
Their choreography groups occur in training with other characters, and $6$ of the $136$ test clips share their motion trajectory with a training clip.
\cref{tab:supp_overlap} compares these $6$ clips with the other $130$: PSNR and LPIPS are better on them and the VLM motion rating is lower, a random set of $6$ test clips reaches their Stage~2 PSNR in $38\%$ of the draws, and removing them changes the averages by at most $0.02$\,dB.

\begin{table}[h]
\centering
\small
\caption{\textbf{Test clips with and without a trajectory in training.} PSNR / LPIPS / VLM motion on the $136$ engine renders.}
\label{tab:supp_overlap}
\setlength{\tabcolsep}{4pt}
\resizebox{\linewidth}{!}{%
\begin{tabular}{@{}lccc@{}}
\toprule
Model & All $136$ & Trajectory in training ($6$) & Other $130$ \\
\midrule
Teacher & 21.21 / 0.143 / 4.50 & 21.36 / 0.116 / 4.17 & 21.21 / 0.144 / 4.52 \\
Stage~1 & 19.82 / 0.187 / 4.45 & 19.90 / 0.160 / 4.17 & 19.81 / 0.188 / 4.47 \\
Stage~2 & 20.10 / 0.186 / 4.49 & 20.38 / 0.155 / 4.17 & 20.09 / 0.187 / 4.50 \\
\bottomrule
\end{tabular}}
\end{table}

\paragraph{Reconstruction metrics}
All outputs are resampled to $16$\,fps and $480{\times}864$ and compared with the original ground-truth frames, not with their VAE reconstruction, over frames $1$ to $80$; external clips shorter than $5$\,s are evaluated over their own length.
SSIM is computed on luminance with an $11{\times}11$ Gaussian window, and LPIPS with the AlexNet variant~\cite{lpips}.
We report means with bootstrap $95\%$ confidence intervals over clips ($5{,}000$ resamples); \cref{tab:supp_full} lists all metrics of the comparison.

\begin{table*}[t]
\centering
\small
\caption{\textbf{All metrics of the comparison} (\cref{tab:main}), with bootstrap $95\%$ confidence intervals over clips for PSNR and the motion rating. Quality and Temp.: ratings of the visual quality and of the consistency of the four frames. Marionette is evaluated on the $80$ external clips for which its control could be built; on the other $28$, no body of positive size is found for its rasterization.}
\label{tab:supp_full}
\setlength{\tabcolsep}{4.5pt}
\begin{tabular}{@{}llcccccccc@{}}
\toprule
Control & Method & MSE$\downarrow$ & PSNR$\uparrow$ & SSIM$\uparrow$ & LPIPS$\downarrow$ & Motion$\uparrow$ & App.$\uparrow$ & Quality$\uparrow$ & Temp.$\uparrow$ \\
\midrule
\multicolumn{10}{@{}l}{\emph{Engine renders ($66$ clips)}} \\
2D pose & Marionette & 0.0447 & 14.49 [13.92, 15.07] & 0.426 & 0.544 & 2.87 [2.78, 2.95] & 2.77 & 2.50 & 2.52 \\
2D pose & Fun-5B-Control & 0.0291 & 16.62 [15.90, 17.36] & 0.441 & 0.402 & 3.55 [3.43, 3.67] & 4.00 & 3.60 & 3.84 \\
Depth & Fun-5B-Control & 0.0309 & 16.46 [15.71, 17.23] & 0.438 & 0.407 & 4.11 [4.02, 4.20] & 4.05 & 3.77 & 4.03 \\
G-buffer & World Renderer & 0.1511 & 9.09 [8.44, 9.79] & 0.224 & 0.851 & 3.81 [3.72, 3.89] & 1.14 & 1.67 & 3.61 \\
3D proxies & Fun-5B-Control & 0.0306 & 16.46 [15.73, 17.22] & 0.439 & 0.407 & 4.28 [4.17, 4.39] & 3.95 & 3.70 & 4.15 \\
3D proxies & Ours, teacher & 0.0098 & 21.31 [20.82, 21.84] & 0.577 & 0.146 & 4.52 [4.41, 4.62] & 4.69 & 3.97 & 4.64 \\
3D proxies & Ours, Stage~1 & 0.0147 & 19.37 [18.78, 19.99] & 0.519 & 0.218 & 4.45 [4.36, 4.56] & 4.58 & 3.92 & 4.63 \\
3D proxies & Ours, Stage~2 & 0.0119 & 20.31 [19.74, 20.94] & 0.558 & 0.184 & 4.51 [4.40, 4.61] & 4.67 & 3.95 & 4.63 \\
3D proxies & \quad first frame only & 0.0130 & 20.04 [19.42, 20.70] & 0.553 & 0.201 & 4.44 [4.34, 4.55] & 4.64 & 3.92 & 4.58 \\
\midrule
\multicolumn{10}{@{}l}{\emph{External clips ($108$ clips)}} \\
2D pose & Marionette ($80$ clips) & 0.0576 & 13.54 [13.05, 14.06] & 0.365 & 0.587 & 2.58 [2.40, 2.75] & 2.81 & 2.16 & 2.08 \\
2D pose & Fun-5B-Control & 0.0410 & 15.46 [14.81, 16.15] & 0.457 & 0.490 & 3.68 [3.56, 3.79] & 4.39 & 3.64 & 3.95 \\
Depth & Fun-5B-Control & 0.0440 & 14.95 [14.34, 15.61] & 0.441 & 0.513 & 3.83 [3.72, 3.94] & 4.24 & 3.47 & 3.92 \\
G-buffer & World Renderer & 0.1341 & 9.87 [9.27, 10.48] & 0.234 & 0.856 & 3.52 [3.36, 3.69] & 1.00 & 1.22 & 3.76 \\
3D proxies & Fun-5B-Control & 0.0436 & 15.04 [14.43, 15.71] & 0.446 & 0.516 & 3.92 [3.81, 4.02] & 4.18 & 3.48 & 3.96 \\
3D proxies & Ours, teacher & 0.0377 & 15.61 [15.01, 16.26] & 0.458 & 0.464 & 3.85 [3.75, 3.94] & 4.69 & 3.75 & 4.08 \\
3D proxies & Ours, Stage~1 & 0.0517 & 14.14 [13.57, 14.73] & 0.427 & 0.589 & 3.70 [3.60, 3.81] & 4.37 & 3.59 & 3.66 \\
3D proxies & Ours, Stage~2 & 0.0690 & 12.79 [12.26, 13.36] & 0.397 & 0.611 & 3.66 [3.55, 3.76] & 4.35 & 3.81 & 3.68 \\
3D proxies & \quad first frame only & 0.0830 & 12.05 [11.53, 12.59] & 0.376 & 0.636 & 3.57 [3.47, 3.69] & 4.02 & 3.76 & 3.39 \\
\bottomrule
\end{tabular}
\end{table*}

\paragraph{Paired comparisons}
The differences below are means of per-clip differences with bootstrap $95\%$ intervals.
On every one of the $66$ engine clips, Stage~2 has a higher PSNR and a lower LPIPS than each of the five baselines, and so has its first-frame-only variant. On average, Stage~2 exceeds the baseline with the best PSNR, Fun-5B-Control with skeletons, by $3.70$\,dB $[3.24, 4.14]$, and the baseline with the best motion rating, Fun-5B-Control with our proxies, by $0.23$ $[0.11, 0.35]$ in this rating.
Driving Fun-5B-Control with our proxies instead of depth changes its motion rating by $+0.17$ $[0.08, 0.27]$ and its PSNR by $+0.00$\,dB $[-0.10, 0.11]$; instead of skeletons, by $+0.73$ $[0.61, 0.85]$ and $-0.16$\,dB $[-0.28, -0.04]$.
On the $108$ external clips, the corresponding changes of the motion rating are $+0.08$ $[-0.01, 0.18]$ against depth and $+0.24$ $[0.13, 0.35]$ against skeletons, and skeletons give $0.41$\,dB $[0.25, 0.60]$ more PSNR than proxies.
Stage~2 differs from Stage~1 by $+0.95$\,dB $[0.74, 1.17]$ on engine renders with the static caption ($+0.25$\,dB $[0.09, 0.40]$ with the clip caption, with which Stage~1 reaches $20.02$\,dB) and by $-1.35$\,dB $[-1.68, -1.00]$ on external clips.

\paragraph{Multimodal rating}
For each clip and method, a panel shows the control and the output at $0.5$, $1.5$, $3.0$ and $4.5$\,s together with thumbnails of the appearance inputs.
The model (GPT-6.1 with medium reasoning effort, queried on 14--16 November 2026) answers in JSON with scores from $1$ to $5$; panels carry no method names, and each panel is rated twice in independent requests, whose scores we average.
The two ratings of a panel are identical in $78\%$ (motion), $74\%$ (appearance), $87\%$ (quality) and $76\%$ (temporal consistency) of the $2{,}400$ short-clip panels of our models and their ablations, with mean absolute differences between $0.13$ and $0.26$.
The prompt for $5$-s clips is:
\begin{quote}\raggedright\footnotesize\ttfamily
You are evaluating a video rendering method. Row 1 shows a motion-control video: untextured white 3D proxies of the characters on a black background, at four time points. Row 2 shows the generated video at the same time points. The small images on the right show the appearance inputs (first frame, character reference, scene reference). \\
Score each item from 1 (very poor) to 5 (excellent): \\
- motion: do the characters, weapons and camera in Row 2 match the poses, positions, contacts and occlusion order in Row 1? \\
- appearance: do the characters and the scene in Row 2 match the appearance inputs? \\
- quality: is Row 2 free of artifacts, distortions and blur? \\
- temporal: are the four frames of Row 2 consistent with each other? \\
Also list the failures you see in Row 2, chosen from: identity\_swap, wrong\_occlusion\_order, missing\_weapon, wrong\_pose, background\_mismatch, distorted\_body, flicker, none. \\
Answer only with JSON: \{"motion": x, "appearance": x, "quality": x, "temporal": x, "failures": [...], "reason": "<one sentence>"\}
\end{quote}
For the windows of one-minute sequences, the motion and temporal items and their JSON fields are removed.
Four frames can miss short events such as a brief identity swap between them, so the ratings complement the reconstruction metrics rather than replace a frame-accurate check of contacts and occlusions.
\cref{tab:supp_failures} reports how often the model names each failure, averaged over the two ratings of a panel.
A rate of zero means that the rater saw no such failure in these panels, not that none occurs in the video.
The pose tag also marks small deviations: Stage~2 clips with the tag still receive a mean motion rating of $4.12$ (those without it $4.86$), whereas under an empty control or the control of another clip the tag is named in $100\%$ and $98\%$ of the ratings and the motion rating falls to $2.09$ and $3.12$, so the rater does detect a control that is not followed.

\begin{table}[h]
\centering
\small
\caption{\textbf{Failures named by the multimodal rater} (fraction of ratings). Top: all methods on the $66$ engine comparison clips, with the captions of \cref{tab:main}. Bottom: test-time variants of Stage~2 on the $136$ engine renders, with the clip caption.}
\label{tab:supp_failures}
\setlength{\tabcolsep}{3pt}
\resizebox{\linewidth}{!}{%
\begin{tabular}{@{}lccccc@{}}
\toprule
Method & Identity swap & Occlusion order & Missing weapon & Wrong pose & Distorted body \\
\midrule
Marionette & 0.06 & 0.00 & 0.84 & 1.00 & 0.90 \\
Fun-5B-Control, skeleton & 0.00 & 0.01 & 0.27 & 0.86 & 0.41 \\
Fun-5B-Control, depth & 0.02 & 0.00 & 0.02 & 0.63 & 0.27 \\
World Renderer & 0.00 & 0.00 & 0.11 & 0.48 & 0.97 \\
Fun-5B-Control, proxies & 0.05 & 0.00 & 0.02 & 0.48 & 0.31 \\
Ours, teacher & 0.00 & 0.00 & 0.01 & 0.34 & 0.09 \\
Ours, Stage~1 & 0.00 & 0.00 & 0.01 & 0.41 & 0.16 \\
Ours, Stage~2 & 0.00 & 0.00 & 0.01 & 0.35 & 0.14 \\
\midrule
Stage~2 ($136$ clips) & 0.00 & 0.00 & 0.01 & 0.37 & 0.12 \\
\quad flat silhouette & 0.00 & 0.00 & 0.00 & 0.47 & 0.10 \\
\quad skeleton (ground-truth frames) & 0.00 & 0.00 & 1.00 & 0.85 & 0.76 \\
\quad control of another clip & 0.08 & 0.04 & 0.06 & 0.98 & 0.51 \\
\quad empty control & 0.04 & 0.03 & 0.71 & 1.00 & 0.33 \\
\quad references only & 0.03 & 0.00 & 0.01 & 0.44 & 0.13 \\
\quad no appearance input & 0.06 & 0.00 & 0.02 & 0.35 & 0.33 \\
\bottomrule
\end{tabular}}
\end{table}

\paragraph{Frame-level pose and identity}
Because the rater sees four frames, we also run DWPose on frames $1$ to $80$ of the $66$ engine comparison clips, for the ground truth and for each output, match persons per frame, and count a visible ground-truth body joint as correct if the matched joint lies within $0.1$ of the diagonal of the person's joint box (PCK); joints of unmatched persons count as wrong.
PCK is $0.957$ $[0.951, 0.962]$ for Stage~2 and $0.892$, $0.861$ and $0.858$ for Fun-5B-Control with skeletons, depth and proxies; Stage~2 exceeds Fun-5B-Control with proxies by $0.099$ $[0.074, 0.125]$, and frames with PCK below $0.5$ make up $0.8\%$ for Stage~2.
For Fun-5B-Control, proxies and depth do not differ ($-0.004$ $[-0.032, 0.025]$) and skeletons are $0.034$ $[0.003, 0.066]$ better than proxies; the skeleton control comes from the same estimator as this metric, and the metric ignores weapons.
With the control of another clip, PCK is $0.381$ against the poses of the original clip and $0.878$ against those of the clip that supplied the control.
For identity, we crop both matched persons and compare DINOv2 features with the two ground-truth crops; a frame counts as swapped if the crossed assignment is more similar. Stage~2 is correct in $99.5\%$ of the frames and no clip has six consecutive swapped frames; Fun-5B-Control is correct in $93\%$, $85\%$ and $87\%$ of the frames (skeletons, depth, proxies), with such runs in $14\%$ to $21\%$ of its clips.
Over all $330$ clip--method pairs, PCK and the motion rating have a Spearman correlation of $0.54$.
Neither measures weapons or contacts, which we evaluate next.

\paragraph{Contact events}
\label{supp:contact}
To test whether a fast weapon or limb is rendered at the right place and frame when the two characters meet, we locate contact events from the control alone, without looking at any output.
The proxy foreground of every frame (brightness above $40$ of $255$, closed with a $5{\times}5$ kernel) is split into connected components; a frame is a contact frame if the second-largest component has less than $40\%$ of its median area over the clip, \ie, the silhouettes of the two characters, weapons included, have merged.
An event starts at the first frame $t_0$ of a run of contact frames if the two characters are separate at $t_0-1$ and the merged component at $t_0$ has at least $80\%$ of their summed area, which excludes a character that leaves the image; we keep events with $4\le t_0\le 76$.
In our clips this is the frame at which a weapon or a limb meets the opponent in the image (\cref{fig:supp_contact}); the two need not touch in 3D, which the engine record below quantifies.
The $66$ comparison clips contain $72$ events in $59$ clips from $29$ choreography groups; $37$ events last a single frame, because the weapon moves on, and the same attack can recur in several groups ($33$ distinct combinations of frame and contact point). We bootstrap over choreography groups.

Around each event we take a window of $128{\times}128$ pixels centered between the closest points of the two characters at $t_0-1$ and the frames $t_0-3$ to $t_0+3$, and report (\cref{tab:supp_contact}):
the PSNR in this window;
the number of events for which, among the offsets $d\in\{-3,\dots,3\}$, the output frames match the ground-truth frames $t+d$ best at $d=0$ (\emph{aligned}); this measures local alignment in time, not whether the strike itself is rendered;
and the \emph{strike correlation}.
For the latter, let $P$ be the pixels of the window that the control covers at $t_0$ but not within three pixels of what it covers at $t_0-2$ or $t_0+2$: the weapon or limb that moves in at this frame, which we call the striking part.
We correlate the change of the output on $P$, $V_{t_0}-\tfrac12(V_{t_0-2}+V_{t_0+2})$, with the same change of the ground truth; this is the cosine $\langle d_V,d_Y\rangle/(\|d_V\|\|d_Y\|+10^{-6})$ of the two changes $d_V$ and $d_Y$ on $P$ (not mean-centered), which is $1$ if the output changes in the same way as the ground truth, up to a positive factor, and $0$ if it does not change at all. Being invariant to scale, it also rewards a faint trace of the right shape and color, so it measures whether the change is in the right place and direction, not whether the part is rendered in full. $69$ events have at least $40$ such pixels with a visible change in the ground truth.

\begin{table}[h]
\centering
\small
\caption{\textbf{Contact events} on the $66$ engine comparison clips ($72$ events; strike measures on $69$). PSNR: in the window around the event, with bootstrap $95\%$ intervals over choreography groups. Aligned: events whose window matches the ground truth best without a time offset. Corr., PSNR$_P$: strike correlation and PSNR on the pixels $P$ of the striking weapon or limb at $t_0$. App.: share of events (\%) in which the strike appears, \ie, its correlation exceeds $0.5$.}
\label{tab:supp_contact}
\setlength{\tabcolsep}{3pt}
\resizebox{\linewidth}{!}{%
\begin{tabular}{@{}llccccc@{}}
\toprule
Control & Method & PSNR$\uparrow$ & Aligned$\uparrow$ & Corr.$\uparrow$ & App.$\uparrow$ & PSNR$_P\uparrow$ \\
\midrule
2D pose & Marionette & 12.39 [11.54, 13.16] & 11 & 0.11 [0.01, 0.22] & 14 & 10.88 \\
2D pose & Fun-5B-Control & 14.87 [13.98, 15.76] & 53 & 0.19 [0.12, 0.25] & 14 & 12.38 \\
Depth & Fun-5B-Control & 14.60 [13.75, 15.46] & 55 & 0.15 [0.11, 0.18] & 6 & 11.50 \\
G-buffer & World Renderer & 6.70 [6.29, 7.13] & 6 & $-$0.17 [$-$0.28, $-$0.05] & 12 & 5.68 \\
3D proxies & Fun-5B-Control & 14.27 [13.50, 15.03] & 56 & 0.11 [0.05, 0.18] & 4 & 11.12 \\
\midrule
3D proxies & Ours, teacher & 18.45 [17.66, 19.21] & 71 & 0.58 [0.53, 0.62] & 65 & 14.22 \\
3D proxies & Ours, Stage~1 & 17.52 [16.70, 18.27] & 72 & 0.58 [0.53, 0.63] & 67 & 13.49 \\
3D proxies & Ours, Stage~2 & 17.92 [17.20, 18.56] & 72 & 0.59 [0.53, 0.64] & 68 & 13.81 \\
3D proxies & \quad clip caption & 17.82 [16.99, 18.59] & 71 & 0.54 [0.49, 0.61] & 67 & 13.77 \\
3D proxies & \quad first frame only & 17.70 [16.82, 18.51] & 70 & 0.53 [0.47, 0.59] & 62 & 13.69 \\
3D proxies & \quad references only & 16.52 [15.74, 17.29] & 70 & 0.57 [0.52, 0.62] & 70 & 13.21 \\
Silhouette & Ours, Stage~2 & 17.67 [16.96, 18.32] & 71 & 0.55 [0.48, 0.62] & 67 & 13.47 \\
\bottomrule
\end{tabular}}
\end{table}

Our models are aligned in $70$ to $72$ of the $72$ events, Fun-5B-Control in $53$ to $56$ with its three controls.
The window PSNR of Stage~2 is $3.7$\,dB $[3.2, 4.2]$ above that of Fun-5B-Control with proxies, and its strike correlation is higher by $0.48$ $[0.38, 0.56]$.
We say that the strike \emph{appears} if this correlation exceeds $0.5$, an operational threshold whose choice we test below; it does not mean that half of the part is drawn. It appears in $68\%$ $[56, 80]$ of the $69$ events for Stage~2 and in $4$ to $14\%$ for the five baselines; the paired differences lie between $+54$ and $+64$ points, with lower interval bounds of at least $+36$. The choice of the threshold does not matter: at $0.3$, $0.4$, $0.6$ and $0.7$ the shares are $84$, $81$, $55$ and $45\%$ for Stage~2 and at most $33$, $25$, $6$ and $6\%$ for the baselines.
Split by the striking part (see the direct check below), the gap holds at the $48$ weapon strikes ($71\%$ against at most $12\%$; strike correlation $0.58$ against $0.08$ to $0.16$ for Fun-5B-Control; aligned in all $48$ events against $37$ to $40$) and at the $24$ contacts of limbs ($62\%$ against at most $38\%$; $0.61$ against $0.18$ to $0.26$; aligned in all $24$ against $15$ to $17$).
At the event frames ($t_0-3$ to $t_0+3$), the body PCK of the frame-level check is $0.914$ for Stage~2 ($0.962$ on the other frames) and $0.817$, $0.777$ and $0.771$ for Fun-5B-Control with skeletons, depth and proxies ($0.894$, $0.863$ and $0.860$ elsewhere); the two identities are correct in $99.3\%$ of the event frames for Stage~2 and in $86\%$, $82\%$ and $83\%$ for Fun-5B-Control.
The three models of our pipeline do not differ in the strike correlation (Stage~2 against the teacher, both with the clip caption: $-0.03$ $[-0.08, 0.01]$).

\noindent\emph{Engine record of the events.} The events are found in the image, so a merge of the two silhouettes need not be a contact in 3D. The clips and events of this record are not those of the comparison above, and no baseline was run on them.
For the $54$ clips that were rendered for the new characters (\cref{supp:newid}; $48$ clips and six calibration clips with $72$ events), we therefore re-simulated the original configurations in the engine, without rendering, and exported for every native frame ($30$\,fps) the unsigned distance between the triangle surfaces of each character's weapon, hands and forearms, and feet and lower legs and those of the opponent's proxy body, together with the engine's animation states, attack outcomes and health values.
The trace of the re-simulation is not byte-identical to that of the original run, so we checked it against the rendered control: the nearest points of these pairs, projected into the control windows of the events, lie within two pixels of a proxy in $94\%$ of $2{,}451$ cases (at least $85\%$ in each of the $38$ clips with events), against $75$ to $79\%$ when the geometry is shifted by one native frame and $40\%$ with the geometry of another clip.

By this record, $47$ of the $72$ events ($65\%$ $[52, 79]$, bootstrap over the pairs of characters) fall into an attack or the reaction to it between $t_0-2$ and $t_0+2$, and $25$ into an approach, in which both characters walk or guard and their silhouettes merge; at $17$ events ($24\%$) the engine registers a hit, with a loss of health, within three frames of $t_0$.
The nearest pair is a weapon and the opponent's body at $93\%$ of the events.
Between $t_0-2$ and $t_0+2$ the surfaces come within $5$, $10$ and $25$\,cm at $24$, $36$ and $51\%$ of the events (median $0.22$\,m), against $6$, $10$ and $20\%$ of all $3{,}942$ windows of five frames in these clips (median $0.58$\,m).
The events thus select near contacts, but at $64\%$ of them the surfaces stay more than $10$\,cm apart: these are contacts in the image, at which a weapon or limb passes the opponent at a distance. Contacts between two weapons are not part of these distances, and the engine itself counts a hit when the swept blade meets a sphere of radius $0.2$\,m around a bone of the opponent; the surfaces are $0.22$\,m apart on average at the frames of its hits.
Conversely, of the $24$ hits that the engine registers in the evaluated frames, $14$ lie in a run of contact frames and $13$ within three frames of its start; the other ten fall into six clips in which our criterion finds no event.
For Stage~2, the strike appears in $59\%$ of the $44$ measurable events in attacks and in $13$ of the $14$ in approaches, and in $62\%$ of the measurable events within $10$\,cm against $71\%$ of the others: the measure does not depend on a contact in 3D, and the attacks are the harder events.
No such record was exported for the clips of the two test characters, which come from the same generator; the shares above therefore describe the kind of events our criterion selects, not the comparison set, and the terms contact event and strike refer to the image throughout. The re-simulation is checked against the image only, so its distances are an approximate record, not a per-frame ground truth of contact.

\begin{figure*}[t]
\centering
\includegraphics[width=\linewidth]{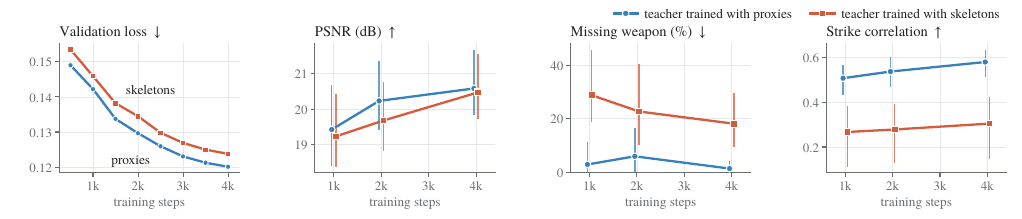}
\caption{\textbf{Proxies and skeletons under one training budget.} Validation loss of the two teachers every $500$ steps (both start at $0.201$), and PSNR, the share of ratings that name a missing weapon and the strike correlation at the contact events after $1{,}000$, $2{,}000$ and $4{,}000$ steps ($33$ test clips). Whiskers: $95\%$ bootstrap intervals of each model over choreography groups; the paired differences are given in the text.}
\label{fig:supp_r5}
\end{figure*}

\noindent\emph{Proxies and skeletons under one budget.} Two teachers are trained with the code and hyperparameters of our teacher from the same initialization (Wan2.2-TI2V-5B) on the same $1{,}000$ engine clips without the test characters (first window, static caption) for $4{,}000$ steps at batch size $4$; they differ only in the control latents, which come from the proxy render or from a skeleton video drawn from DWPose keypoints of the ground-truth frames, at training and at test time. They are sampled like the teacher, with the final weights instead of their average, on the first $33$ of the $66$ comparison clips ($36$ contact events), on other hardware than \cref{tab:main}, so the values are comparable within \cref{tab:supp_r5} only.
The PSNR of the two models does not differ after $4{,}000$ steps ($+0.11$\,dB $[-0.10, 0.32]$), but with proxies the motion rating is higher ($+0.24$ $[0.08, 0.41]$), a missing weapon is named in $2\%$ instead of $18\%$ of the ratings, and at the contact events the window PSNR is higher by $0.39$\,dB $[0.24, 0.71]$ and the strike correlation by $0.28$ $[0.18, 0.41]$, so that the strike appears in $73\%$ $[54, 89]$ of the $33$ events instead of $30\%$ $[9, 47]$ ($+42$ points $[27, 59]$; $67$ vs.\ $30\%$ and $52$ vs.\ $18\%$ after $2{,}000$ and $1{,}000$ steps); these differences have the same sign and a similar size after $1{,}000$ and $2{,}000$ steps (\cref{tab:supp_r5,fig:supp_r5}). The validation loss of the proxy model is lower at every $500$ steps ($0.120$ vs.\ $0.124$ at the end). This is one training run per control at a reduced budget. The right half of the qualitative figure of the paper shows windows of both teachers at four contact events; they are the three events of \cref{fig:supp_contact} and the next event under the same rule, which visits the environments in turn and takes clips in key order, so that no output was inspected for the choice; the third event is a contact of limbs.

\begin{table}[h]
\centering
\small
\caption{\textbf{Proxies and skeletons under one training budget} ($33$ clips, $36$ contact events). Weapon: share of ratings that name a missing weapon. Window, Corr.: window PSNR and strike correlation at the contact events.}
\label{tab:supp_r5}
\setlength{\tabcolsep}{3pt}
\resizebox{\linewidth}{!}{%
\begin{tabular}{@{}llccccccc@{}}
\toprule
Steps & Control & PSNR$\uparrow$ & LPIPS$\downarrow$ & Motion$\uparrow$ & App.$\uparrow$ & Weapon$\downarrow$ & Window$\uparrow$ & Corr.$\uparrow$ \\
\midrule
1{,}000 & Skeleton & 19.22 & 0.248 & 4.00 & 4.36 & 0.29 & 16.90 & 0.27 \\
1{,}000 & Proxies & 19.42 & 0.245 & 4.47 & 4.61 & 0.03 & 17.39 & 0.51 \\
2{,}000 & Skeleton & 19.67 & 0.207 & 4.14 & 4.50 & 0.23 & 17.41 & 0.28 \\
2{,}000 & Proxies & 20.23 & 0.194 & 4.45 & 4.62 & 0.06 & 17.94 & 0.54 \\
4{,}000 & Skeleton & 20.47 & 0.171 & 4.39 & 4.55 & 0.18 & 17.70 & 0.30 \\
4{,}000 & Proxies & 20.59 & 0.173 & 4.64 & 4.71 & 0.02 & 18.09 & 0.58 \\
\bottomrule
\end{tabular}}
\end{table}

Three results limit what the control representation explains beyond this comparison.
For our model, a flat silhouette gives the same values as the shaded proxies with the same caption (window PSNR $-0.15$\,dB $[-0.29, 0.04]$, strike correlation $+0.00$ $[-0.03, 0.03]$), so the shading brings no measurable benefit at these events. Over whole clips the two controls differ: with the silhouette the rater names a wrong pose in $47\%$ of the ratings against $37\%$ (\cref{tab:supp_failures}), and in the videos we inspected weapons lose their shape during strikes, although the direct check below finds no difference between the two controls at the contact frame.
For Fun-5B-Control, proxies are not better than skeletons at these events (window PSNR $-0.60$\,dB $[-0.82, -0.37]$, strike correlation $-0.08$ $[-0.12, -0.03]$): it renders the strike with none of its controls.
And all methods render the pixels of the strike with low fidelity (PSNR$_P$ below $15$\,dB): our models draw the weapon at the frame and place of the control, but darker and with less detail than the ground truth, and a weapon that is visible for a single frame leaves only a faint trace (\cref{fig:supp_contact}).
The alignment measure depends on appearance as well as on timing: the proxy video itself, which is synchronous by construction, is aligned in only $13$ events when compared with the ground truth in the same way. It therefore separates methods whose frames resemble the ground truth, and for Marionette and World Renderer it mainly reflects their appearance.
The measures cover strikes and crossings of the two silhouettes; they do not determine which character is in front.

\noindent\emph{Direct check.} For every event and method, the rater sees five consecutive frames of the control and of the output around $t_0$, without the name of the method, names the part that touches the opponent in the control, and states whether this part is visible in the output at $t_0$ and reaches the opponent at about the same place (yes; partly, \ie, faint, misshapen or displaced; no, \ie, absent or elsewhere), in two independent requests ($144$ answers per method; \cref{tab:supp_direct}). World Renderer, which renders untextured figures from our inputs, was not included.
The two requests give the same answer for $81\%$ of the $504$ panels (Cohen's $\kappa=0.71$) and opposite answers, yes and no, for $0.6\%$; they name the same part in $96\%$.
Over all events, the part is named as first reaching the opponent at $t_0$ in $81\%$ of the answers for the ground truth and $50\%$ for Stage~2, and in none of the five frames in $6\%$ and $29\%$.

\begin{table}[h]
\centering
\small
\caption{\textbf{Direct check at the contact events}: answers yes / partly / no (\%) to whether the striking part is visible at $t_0$ and reaches the opponent, for all $72$ events and split by the part that the rater names in the control.}
\label{tab:supp_direct}
\setlength{\tabcolsep}{3pt}
\resizebox{\linewidth}{!}{%
\begin{tabular}{@{}lccc@{}}
\toprule
Method & All ($72$) & Weapon ($48$) & Limb ($24$) \\
\midrule
Ground truth & 76 / 21 / 3 & 91 / 9 / 0 & 48 / 44 / 8 \\
\midrule
Marionette & 17 / 23 / 60 & 1 / 15 / 84 & 50 / 40 / 10 \\
Fun-5B-Control, skeleton & 20 / 37 / 43 & 4 / 38 / 58 & 52 / 35 / 13 \\
Fun-5B-Control, depth & 17 / 53 / 30 & 9 / 53 / 38 & 33 / 52 / 15 \\
Fun-5B-Control, proxies & 24 / 46 / 30 & 15 / 44 / 42 & 44 / 50 / 6 \\
\midrule
Ours, Stage~2 & 33 / 50 / 17 & 36 / 47 / 17 & 25 / 56 / 19 \\
\quad flat silhouette & 33 / 49 / 18 & 40 / 39 / 22 & 21 / 69 / 10 \\
\bottomrule
\end{tabular}}
\end{table}

The named part is a weapon in $48$ events and an arm, hand, leg or foot in $24$ (majority of the $14$ answers of an event, which are unanimous in $62$ events), and the two kinds differ.
Where a weapon strikes, the check is reliable: the ground truth is judged yes in $91\%$ of the answers and never no, and the two requests agree for $85\%$ of the panels ($\kappa=0.78$). There the weapon is missing (answer no) in $17\%$ of the answers for Stage~2, against $38$ to $58\%$ for Fun-5B-Control and $84\%$ for Marionette, and fully visible (yes) in $36\%$, against $4$ to $15\%$ and $1\%$; against Fun-5B-Control with proxies, the paired differences are $+22$ points $[9, 34]$ for yes and $-25$ points $[-40, -11]$ for no (bootstrap over choreography groups). The weapon is named as first reaching the opponent at $t_0$ in $59\%$ of the answers for Stage~2 ($95\%$ for the ground truth, $18$ to $47\%$ for Fun-5B-Control) and in none of the five frames in $32\%$ ($5\%$; $34$ to $55\%$).
At contacts of limbs it is less reliable: the ground truth itself is judged yes in only $48\%$ of the answers, and the two requests agree for $71\%$ of the panels ($\kappa=0.52$). There it does not favor our model: yes in $25\%$, against $33$ to $52\%$ for Fun-5B-Control and $50\%$ for Marionette ($-27$ points $[-58, -5]$ against Fun-5B-Control with skeletons), whereas the pixel measures above do.
The answers of the rater and the pixel measure are related but not redundant: at weapon strikes, the mean strike correlation is $0.45$ for panels answered yes, $0.27$ for partly and $0.18$ for no (Spearman correlation $0.29$ over the $288$ panels of the six methods).
Our model thus keeps a striking weapon far more often than the baselines but renders it fully in only a third of the events, far below the ground truth; at contacts of limbs the rater and the pixel measures disagree, which the human check below resolves against the rater.
\noindent\emph{Human check.} Ten volunteers who are not authors answered the question of the direct check (yes, partly or no) for the $288$ panels of the ground truth, Stage~2 and Fun-5B-Control with proxies and with skeletons at all $72$ events. The panels carried no method names; the first $60$ were shown in one fixed order and the others in one random order. Each volunteer answered alone and did not see the answers of the others. An automated tool then entered all answers into the result files in one pass, so that the files share their time stamps and response times are not meaningful; the authors neither answered nor entered answers. Intervals below are bootstrapped over choreography groups on the mean answer of the ten volunteers per panel; the volunteers themselves are not resampled. Two volunteers give the same answer for $84\%$ of the panels on average ($76$ to $92\%$; Fleiss' $\kappa=0.74$; $0.77$ at weapon strikes and $0.56$ at contacts of limbs).
\cref{tab:supp_human} gives the shares over all answers.
At weapon strikes the volunteers confirm the rater: the weapon is missing in $16\%$ $[10, 23]$ of the answers for Stage~2 and in $51\%$ and $65\%$ for Fun-5B-Control with proxies and skeletons ($-34$ points $[-50, -23]$ and $-49$ points $[-61, -36]$), and fully visible in $31\%$ against $12\%$ and $11\%$ ($+18$ points $[11, 27]$ and $+19$ points $[9, 29]$), far below the ground truth ($95\%$). A volunteer and the rater give the same answer for $63\%$ of these panels ($\kappa=0.45$ on average) and opposite answers for $3\%$.
At contacts of limbs they do not confirm the rater ($\kappa=0.02$): the volunteers judge the ground truth fully visible in $95\%$ of the answers, not $48\%$, and Stage~2 in $84\%$, against $63\%$ and $54\%$ for Fun-5B-Control ($+20$ points $[4, 43]$ and $+30$ points $[17, 44]$). The answers of the rater at contacts of limbs should therefore not be used; those of the volunteers agree with the pixel measures.
The panels, the ten original answer files of the volunteers and the answers of the rater will be released with the data.

\begin{table}[h]
\centering
\small
\caption{\textbf{Human check at the contact events}: answers yes / partly / no (\%) of ten volunteers, next to those of the rater (\cref{tab:supp_direct}).}
\label{tab:supp_human}
\setlength{\tabcolsep}{3pt}
\resizebox{\linewidth}{!}{%
\begin{tabular}{@{}llcc@{}}
\toprule
Events & Method & Volunteers & Rater \\
\midrule
Weapon ($48$) & Ground truth & 95 / 5 / 0 & 91 / 9 / 0 \\
 & Fun-5B-Control, skeleton & 11 / 24 / 65 & 4 / 38 / 58 \\
 & Fun-5B-Control, proxies & 12 / 37 / 51 & 15 / 44 / 42 \\
 & Ours, Stage~2 & 31 / 53 / 16 & 36 / 47 / 17 \\
\midrule
Limb ($24$) & Ground truth & 95 / 1 / 4 & 48 / 44 / 8 \\
 & Fun-5B-Control, skeleton & 54 / 33 / 13 & 52 / 35 / 13 \\
 & Fun-5B-Control, proxies & 63 / 29 / 8 & 44 / 50 / 6 \\
 & Ours, Stage~2 & 84 / 14 / 2 & 25 / 56 / 19 \\
\bottomrule
\end{tabular}}
\end{table}

The row of Stage~2 uses the static caption, as in \cref{tab:main}; its variants (first frame only, references only, silhouette) and the frame-level PCK and identity checks were run with the clip caption, with which Stage~2 gives the row below it.

\paragraph{Text prompt}
Every engine clip has two captions from one annotation (\cref{supp:data}): the clip caption, which describes the motion of the whole clip, and a static caption, which names the scene and the characters but no motion.
A streaming renderer does not know the motion in advance, so the causal students receive the static caption in the comparison of \cref{tab:main}; the teacher and the baselines, which see the whole control anyway, receive the clip caption, and all methods receive the same generic sentence on external clips, which have no captions.
The inference-time ablations, the one-minute rollouts with other histories and of Stage~1 and the frame-level pose and identity checks were run with the clip caption; the contact measures and the one-minute results of the main paper use the static caption.
The caption makes no difference: with the clip caption, Stage~2 reaches PSNR $20.27$, LPIPS $0.188$, motion $4.50$ and appearance $4.58$ on the $66$ comparison clips, against $20.31$, $0.184$, $4.51$ and $4.67$ with the static caption; the paired differences are $-0.04$\,dB $[-0.21, 0.12]$ and $-0.01$ $[-0.12, 0.11]$ in the motion rating.
The image strips of \cref{fig:qualitative} (left), \cref{fig:supp_comparison}, \cref{fig:supp_contact}, \cref{fig:drift} and \cref{fig:supp_ablation}, the curves of \cref{fig:drift} and the engine test clips of the supplementary video show the runs with the clip caption; the contact windows of \cref{fig:qualitative} (right) and all clips of new characters, in \cref{fig:teaser} and in the video, show Stage~2 with the static caption.
With one generic prompt for all clips, PSNR falls by $4.2$\,dB $[3.7, 4.7]$ and the motion rating by $0.18$ $[0.05, 0.32]$: in the outputs we inspected, the characters still follow the control, but the background drifts as if the camera moved.
The proxies carry no scene geometry, so without a description of the scene the model is free to move it, and a deployment should keep a static scene description.

\paragraph{One-minute sequences}
The control and target of a test clip are played forward and backward in alternation until they reach $969$ frames ($60.6$\,s, $243$ latent frames), and Stage~2 generates the sequence in one rollout without resetting its cache.
Metrics are computed on $[0,10)$\,s and $[50,60)$\,s against the equally extended ground truth. The ratings of these windows use the prompt without the motion and temporal items; a separate rating of the $19$ default rollouts with the full prompt gives motion $4.47$ / $4.74$, appearance $4.66$ / $4.21$, quality $3.97$ / $3.89$ and temporal consistency $4.61$ / $4.76$ for the first / last $10$\,s; DINO is the cosine similarity of DINOv2~\cite{dinov2} ViT-L/14 class tokens between generated frames and the first frame.
The rollout without references removes both reference images, so that nothing anchors the appearance once the first frame has left the cache; we did not test keeping the first frame in the cache instead, so the comparison shows that a persistent anchor is needed and that the references provide one, not that references are better than a retained first frame.
With one denoising step, the setting of the pipeline on three GPUs, whose output equals that of one GPU, the rollouts keep their fidelity (last row of \cref{tab:supp_history}): the late PSNR is $0.26$\,dB $[0.08, 0.43]$ above that with four steps, and LPIPS and the DINO similarity are not worse. Ratings were returned for $17$ of these $19$ rollouts; on them, the late appearance rating does not differ from four steps ($-0.06$ $[-0.38, 0.26]$), and the late quality rating is lower by $0.32$ $[0.06, 0.59]$.

\noindent\emph{One step at the contact events.} We also ran one denoising step with the static caption on the $66$ engine clips of \cref{tab:supp_contact} and repeated the measures of this section. Against four steps on the same clips, PSNR is $20.23$ against $20.32$\,dB ($-0.09$ $[-0.20, 0.02]$), LPIPS $0.187$ against $0.184$, and the ratings of quality and temporal consistency are lower by $0.09$ $[0.03, 0.15]$ and $0.21$ $[0.10, 0.32]$. The strike appears in $67\%$ $[55, 78]$ of the $69$ measurable events, against $68\%$ (paired difference $-1$ point $[-8, 5]$; $88$, $81$, $57$ and $42\%$ at the thresholds $0.3$, $0.4$, $0.6$ and $0.7$). In the direct check, the rater finds the striking part of a weapon fully visible, partly visible and missing in $30$, $64$ and $6\%$ of the answers ($36$, $47$ and $17\%$ with four steps; difference of missing $-10$ points $[-23, 0]$) and that of a limb in $23$, $69$ and $8\%$ ($25$, $56$ and $19\%$). By these automatic measures one step keeps most of the performance at the events, but the interval of the paired difference still allows a small loss, the temporal rating is lower, and the volunteers did not see these panels, so we do not claim that one and four steps are equivalent there.

\paragraph{Stage~1 with the same cache}
Run with the same cache, sink and history as Stage~2 (its own four flow-map steps), Stage~1 takes $16.6$\,s for $5$\,s of video ($4.6$ frames per second), the same as Stage~2.
On the $19$ one-minute sequences it reaches PSNR $17.70$ / $12.90$, LPIPS $0.263$ / $0.518$, DINO $0.880$ / $0.805$, appearance $4.42$ / $3.47$ and quality $3.87$ / $3.34$ in the first / last $10$\,s, against $18.32$ / $14.88$, $0.251$ / $0.450$, $0.885$ / $0.854$, $4.53$ / $4.05$ and $3.92$ / $3.82$ for Stage~2.

\paragraph{Clips used for checkpoint selection}
All results in the paper exclude the selection clips listed above.
For completeness, \cref{tab:supp_selection} reports our models on the $64$ excluded engine renders.

\begin{table}[h]
\centering
\small
\caption{\textbf{Excluded selection clips.} PSNR / LPIPS of our models on the $64$ engine renders used for checkpoint selection and on the $136$ engine renders of the test set.}
\label{tab:supp_selection}
\setlength{\tabcolsep}{5pt}
\begin{tabular}{@{}lcc@{}}
\toprule
Model & Selection $64$ (excluded) & Test $136$ \\
\midrule
Teacher & 20.72 / 0.145 & 21.21 / 0.143 \\
Stage~1 & 19.33 / 0.193 & 19.82 / 0.187 \\
Stage~2 & 19.59 / 0.192 & 20.10 / 0.186 \\
\bottomrule
\end{tabular}
\end{table}

\paragraph{Streaming benchmark}
Control is encoded and video decoded chunk by chunk with the causal VAE, keeping its feature caches across chunks.
Timings use CUDA events after three warm-up chunks and average chunks $10$ to $40$, on an NVIDIA A100-SXM4 with $80$\,GB, PyTorch~2.6 with CUDA~12.4 and its scaled-dot-product attention, without torch.compile.
Chunked encoding, generation and decoding reproduce the output of a single pass over the clip exactly.
Throughput counts encoding, generation, decoding and the copy to the host, but not the writing of evaluation videos.
The cold time to first frame includes the text encoder, the encoding of the first frame and the references and the commit of the reference chunk; the preloaded time starts after these steps.
For the switch latency, the control switches to that of another clip at a chunk boundary, and we measure the time from the arrival of the new control to the first decoded frame that follows it; in all five trials the frames before the switch are bit-identical to the unchanged rollout (maximum absolute difference $0$), and the first changed frame is the first frame of the switched chunk. The bound $0.75\,\text{s}+t_{\text{chunk}}$ of the main paper assumes that generation keeps up with real time and that control is handed over per chunk without queuing; below real time, the engine has to run at the pace of the renderer, and an input that arrives during a chunk waits up to $t_{\text{chunk}}$ for the next chunk boundary, so it appears within $2\,t_{\text{chunk}}$.
Compiling the transformer or the VAE with torch.compile speeds up Stage~2 by $12\%$ and $9\%$, and FlashAttention-2 by $1\%$, but each changes the output at the level of floating-point rounding; all reported numbers therefore use the uncompiled model, with an asynchronous copy to the host that leaves the output bit-identical.
\cref{tab:profile} breaks down the cost of a chunk.

\paragraph{Pipeline on several GPUs}
A block must wait for the cache commit of the previous one, so the transformer cannot be split across GPUs by block, but control encoding and decoding can overlap with it.
With one denoising step, the three stages on three A100s take $0.22$\,s (control encoding), $0.35+0.36$\,s (denoising and commit) and $0.69$\,s (decoding) per chunk, and frames arrive every $0.69$\,s: $17.3$ frames per second on three samples ($17.2$ to $17.4$), above the $16$ frames per second of the video.
Latents and decoded frames are bit-identical to the single-GPU run with one step, whose quality is in \cref{tab:supp_speed} and whose one-minute rollouts are in \cref{tab:supp_history}.
The first frame appears after $1.2$\,s with preloaded inputs and $1.6$\,s otherwise.
Two-slot queues between the stages keep the pipeline busy but delay a control switch to $3.7$\,s ($3.6$ to $3.7$ in five trials, frames before the switch bit-identical): the three stages sum to $1.6$\,s per chunk, and the other $2.1$\,s are waiting time in the queues, which we did not tune.
On two GPUs, with encoding and decoding sharing one, the same setting reaches $14.0$ frames per second.
The default four steps cannot reach real time in this layout, since the transformer alone takes $1.8$\,s per chunk.
\cref{tab:supp_speed} trades speed against quality with fewer denoising steps and a shorter history.

\begin{table}[h]
\centering
\small
\caption{\textbf{Inference cost} of Stage~2 on one A100 at $480{\times}864$ in steady state, without compilation. A chunk is $12$ frames ($0.75$\,s); real time requires $16$ frames per second.}
\label{tab:profile}
\setlength{\tabcolsep}{4pt}
\begin{tabular}{@{}lr@{\hspace{12pt}}lr@{}}
\toprule
\multicolumn{2}{@{}l}{Per chunk (ms)} & \multicolumn{2}{l@{}}{Streaming} \\
\midrule
Control encoding & 169 & Frames/s, 1 GPU & 4.5 \\
Denoising, 4 passes & 1414 & \quad 1 step, 3 GPUs & 17.3 \\
Cache commit & 366 & First frame (ms) & 1844 \\
Decoding & 717 & \quad cold start (ms) & 2162 \\
Copy to host & 2.5 & Switch latency (ms) & 2686 \\
Total, $t_{\text{chunk}}$ & 2669 & Peak memory (GB) & 38.1 \\
\bottomrule
\end{tabular}
\end{table}

\begin{table}[h]
\centering
\small
\caption{\textbf{Speed and quality} of Stage~2 on one A100. Quality: PSNR and LPIPS on the $136$ engine renders of the test set for the steps, and in the last $10$\,s of the $19$ one-minute rollouts for the history.}
\label{tab:supp_speed}
\setlength{\tabcolsep}{6pt}
\begin{tabular}{@{}lccc@{}}
\toprule
Setting & Frames/s & PSNR$\uparrow$ & LPIPS$\downarrow$ \\
\midrule
1 step & 7.5 & 20.05 & 0.186 \\
2 steps & 6.2 & 20.23 & 0.180 \\
4 steps (default) & 4.5 & 20.10 & 0.186 \\
\midrule
History of 3 latents & 5.5 & 14.98 & 0.457 \\
History of 9 latents & 5.0 & 15.00 & 0.452 \\
History of 15 latents (default) & 4.5 & 14.88 & 0.450 \\
\bottomrule
\end{tabular}
\end{table}

\section{Baseline protocols}
\label{supp:baselines}
Engine depth, normals and joints were not exported for our clips, so all controls are computed from the proxy video or, for skeletons, from the ground-truth frames. None of the baselines accepts character or scene references; each receives the first frame $I$.
Skeletons are DWPose~\cite{dwpose} whole-body keypoints estimated on the ground-truth frames: on the proxies, DWPose detects both fighters in fewer than $7\%$ of the frames of $20$ development clips and in $0.1\%$ of the frames of $100$ test clips, so skeletons from the proxies would leave the skeleton-driven methods almost without control.
Depth is the relative depth of Video Depth Anything~\cite{videodepthanything}, normalized per clip between its 1st and 99th percentile over the proxy pixels, with near surfaces bright and the background black.
Normals follow from this depth by Sobel gradients in camera coordinates, and the G-buffers combine depth and normals with a constant white albedo and constant material values.
Marionette takes an RGB control video in which every bone is rendered as a depth-tested cylinder whose red channel encodes the height above the ground, whose green channel encodes the entity and joint index and whose blue channel encodes the inverse depth (checked against its rasterization code).
We build this video with its own shader from the DWPose keypoints of the ground-truth frames: the two longest tracks are mapped to its two character skeletons from left to right, the height of a joint is measured from the lowest foot keypoint under an assumed body height of $1.75$\,m, and its depth is sampled from the estimated depth of the proxy video, both rescaled to the ranges of its calibration data.
This approximates the engine states it was trained with: the camera, the metric scale and the height are reconstructed from 2D keypoints, and its weapon and terrain entities are not produced, so the control contains no weapons.
For $28$ external clips no body of positive size is found and no control is built.
It receives the first frame and the clip caption, and its default negative prompt is kept.
For the G-buffer renderer we evaluate the offline model of Generative World Renderer~\cite{genworldrenderer} with its released weights, sampled with $50$ steps and guidance $5$; it takes G-buffers and text and no image. Its few-step streaming successor~\cite{genworldrendererfast} is listed in \cref{tab:position} but was not evaluated.
Marionette is sampled with $40$ steps and guidance $6$ in one pass over the $81$ frames, and Fun-5B-Control with $50$ steps and guidance $6$; all three use classifier-free guidance, hence two transformer evaluations per step in \cref{tab:main}.
Magpie~\cite{magpie} and Wizard~\cite{wizard} had released no weights when we ran the comparison, and the released SolarRenderer~\cite{solarrenderer} consists of adapters for a backbone several times larger than the $5$B models compared here.

\paragraph{Checkpoints}
For our models we evaluate the EMA weights of the teacher after $6{,}000$ steps, of Stage~1 after $3{,}000$ steps and of Stage~2 after $2{,}000$ iterations; the SHA-256 digests of the three files begin with \texttt{3e10ce9c}, \texttt{e2801583} and \texttt{b5e98af3}.
World Renderer is run from its public repository at commit \texttt{38351f63} and Marionette from its repository at commit \texttt{7c38ddde}, both with the released weights; Fun-5B-Control is revision \texttt{b8bc1a65} of its model repository, run with its control-video and start-image inputs, without a reference image, and with its TeaCache acceleration (threshold $0.1$), which is included in its time.
All three receive the clip caption as prompt.

\section{Additional results}

\paragraph{Settings of related systems}
\cref{tab:position} compares the control and appearance inputs of related systems.

\begin{table}[h]
\centering
\small
\caption{\textbf{Settings of related systems.} Control: the signal that drives the content. Appearance: what fixes the look. Causal: frames are generated from past and present control only.}
\label{tab:position}
\setlength{\tabcolsep}{3pt}
\resizebox{\linewidth}{!}{%
\begin{tabular}{@{}llll@{}}
\toprule
Method & Control & Appearance & Causal \\
\midrule
GameNGen~\cite{gamengen} & player actions & learned, one game & \cmark \\
Matrix-Game 2.0~\cite{matrixgame2} & keyboard, mouse & first frame & \cmark \\
WanToFight~\cite{wantofight} & keyboard, two players & first frame, references & \cmark \\
\midrule
FrameDiffuser~\cite{framediffuser} & G-buffer & engine assets & \cmark \\
World Renderer~\cite{genworldrenderer} & G-buffer & engine assets & \xmark \\
\quad at the speed of play~\cite{genworldrendererfast} & G-buffer & engine assets & \cmark \\
NeuralSRender~\cite{neuralsrender} & G-buffer, text & engine assets & \xmark \\
Marionette~\cite{marionette} & rasterized pose & first frame & \xmark \\
SolarRenderer~\cite{solarrenderer} & structure video & reference (opt.) & \cmark \\
Magpie~\cite{magpie} & untextured frames & first frame & \cmark \\
Wizard~\cite{wizard} & geometry, skeletons & first frame, references & \cmark \\
\midrule
Wan-Animate~\cite{wananimate} & skeleton, face & reference image & \xmark \\
MotionStream~\cite{motionstream} & point tracks, text & first frame & \cmark \\
\midrule
\method & 3D proxies & first frame, references & \cmark \\
\bottomrule
\end{tabular}}
\end{table}

\providecommand{\xrow}[3][]{\rowlab[#1]{#2}\hspace{3pt}\raisebox{-0.5\height}{\includegraphics[width=0.895\linewidth]{fig/img/x_#3.jpg}}\\[1.5pt]}
\providecommand{\xhead}[1]{\makebox[\dimexpr0.895\linewidth*964/3898\relax]{\scriptsize #1}}
\providecommand{\xgap}{\hspace{\dimexpr0.895\linewidth*14/3898\relax}}
\begin{figure*}[t]
\centering
\rowlab{\mbox{}}\hspace{3pt}\xhead{MultiCamVideo}\xgap\xhead{MIND}\xgap\xhead{MiraData}\xgap\xhead{OmniWorld}\\[1pt]
\xrow[3D proxies]{Control}{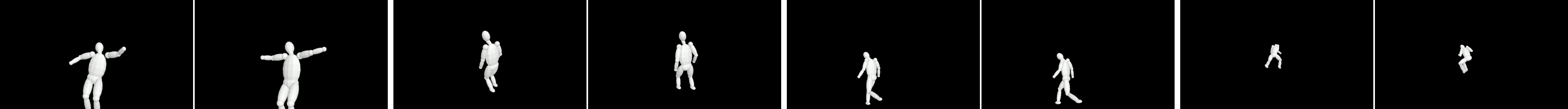}
\xrow[depth]{Fun-5B}{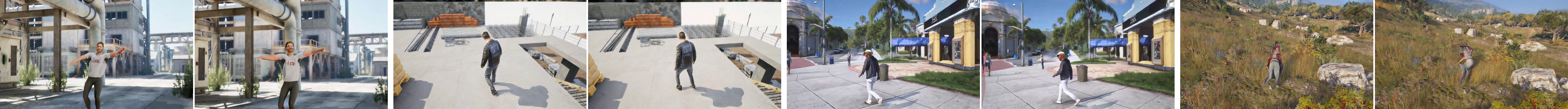}
\xrow[teacher]{\ours}{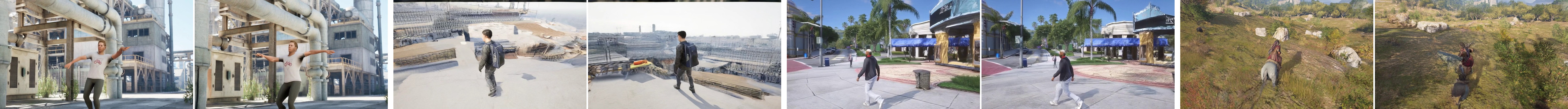}
\xrow[Stage~2]{\ours}{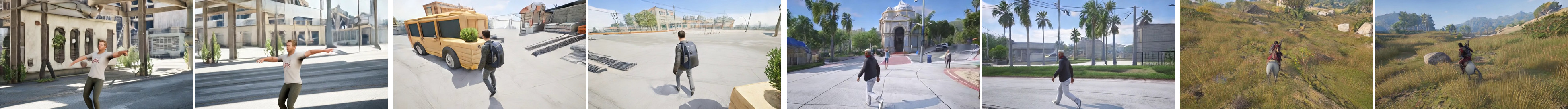}
\xrow{Ground truth}{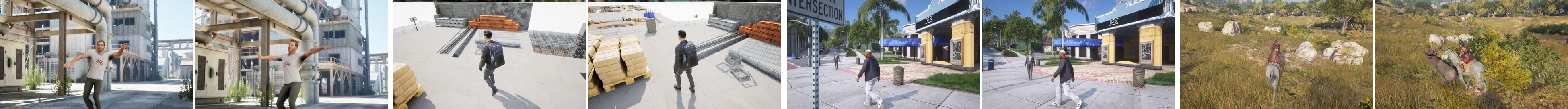}
\caption{\textbf{External clips}, which lie outside our target task: one clip per dataset at $1.0$ and $3.0$\,s. The proxies are reconstructed from the video and cover only the person (the horse of the OmniWorld clip has no proxy), and the camera moves. In these examples all models keep the person on the proxy and differ in the scene: Fun-5B-Control and our teacher stay close to the first frame, whereas Stage~2 moves and changes the scene, which lowers its PSNR (\cref{tab:supp_sources}).}
\label{fig:results}
\end{figure*}

\label{supp:results}
\paragraph{History length on one-minute sequences}
\cref{tab:supp_history} varies the number of history latent frames $h$ kept in the cache.
A shorter history raises the throughput and leaves the reconstruction metrics of the last $10$\,s unchanged, but lowers the VLM ratings in the first $10$\,s and the appearance rating in the last $10$\,s for $h{=}9$.

\begin{table}[h]
\centering
\small
\caption{\textbf{History length, caption and steps} on one-minute rollouts of Stage~2 ($19$ engine renders). Each cell: first $10$\,s / last $10$\,s. Throughput on one A100. $^\dagger$Ratings returned for $17$ of the $19$ rollouts.}
\label{tab:supp_history}
\setlength{\tabcolsep}{3pt}
\resizebox{\linewidth}{!}{%
\begin{tabular}{@{}lcccccc@{}}
\toprule
$h$ & PSNR$\uparrow$ & LPIPS$\downarrow$ & DINO$\uparrow$ & VLM qual.$\uparrow$ & VLM app.$\uparrow$ & Frames/s \\
\midrule
$3$ & 18.04 / 14.98 & 0.260 / 0.457 & 0.879 / 0.857 & 3.58 / 3.82 & 4.13 / 4.00 & 5.55 \\
$9$ & 18.27 / 15.00 & 0.252 / 0.452 & 0.883 / 0.855 & 3.71 / 3.84 & 4.29 / 3.79 & 4.98 \\
\rowcolor{linkblue!12} $15$ & 18.32 / 14.88 & 0.251 / 0.450 & 0.885 / 0.854 & 3.92 / 3.82 & 4.53 / 4.05 & 4.50 \\
\midrule
$15$, no references & 18.03 / 11.44 & 0.265 / 0.601 & 0.884 / 0.762 & 3.84 / 3.00 & 4.34 / 2.26 & 4.50 \\
\midrule
$15$, static caption & 18.39 / 15.33 & 0.246 / 0.431 & 0.888 / 0.861 & 3.61 / 3.82 & 4.45 / 4.32 & 4.50 \\
$15$, static caption, no references & 18.06 / 11.93 & 0.263 / 0.588 & 0.887 / 0.791 & 3.79 / 3.08 & 4.32 / 2.61 & 4.50 \\
$15$, static caption, $1$ step & 18.33 / 15.59 & 0.247 / 0.411 & 0.885 / 0.859 & 3.68 / 3.56$^\dagger$ & 4.41 / 4.29$^\dagger$ & 7.50 \\
\bottomrule
\end{tabular}}
\end{table}

\paragraph{Comparison with all baselines}
\cref{fig:supp_comparison} adds the baselines of \cref{tab:main} that \cref{fig:qualitative} leaves out, and our teacher and Stage~1.
\providecommand{\qstrip}[2]{\raisebox{-0.5\height}{\includegraphics[width=0.445\linewidth]{fig/img/q_#1_#2.jpg}}}
\providecommand{\qrow}[3][]{\rowlab[#1]{#2}\hspace{3pt}\qstrip{#3}{A}\hspace{4pt}\qstrip{#3}{B}\\[1.5pt]}
\providecommand{\qtime}[1]{\makebox[\dimexpr0.445\linewidth*640/2578\relax]{\scriptsize #1}}
\providecommand{\qgap}{\hspace{\dimexpr0.445\linewidth*6/2578\relax}}
\providecommand{\qheads}{\qtime{$0.5$\,s}\qgap\qtime{$1.5$\,s}\qgap\qtime{$3.0$\,s}\qgap\qtime{$4.5$\,s}}
\begin{figure*}[t]
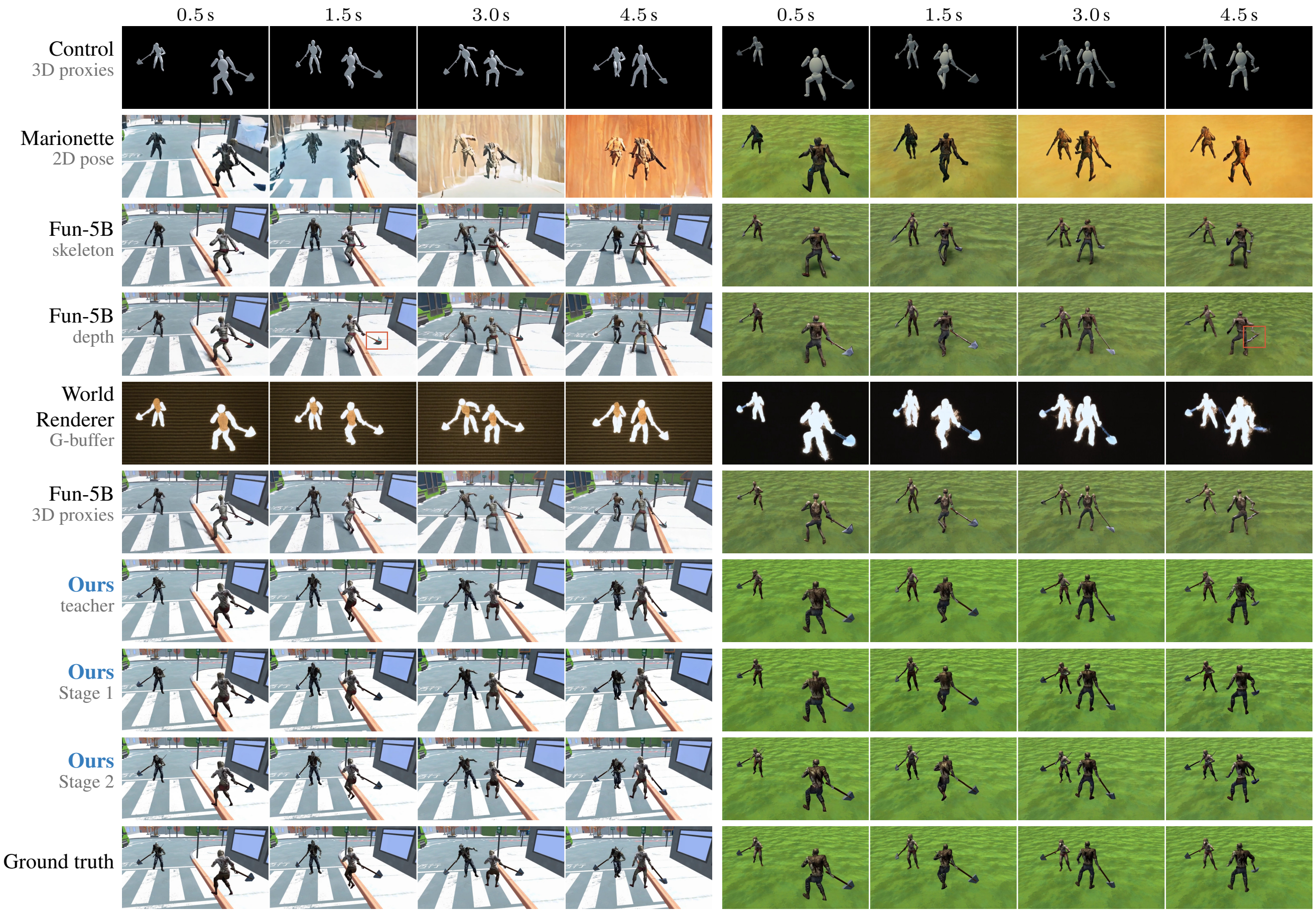

\centering
\rowlab{\mbox{}}\hspace{3pt}\qheads\hspace{4pt}\qheads\\[1pt]
\qrow[3D proxies]{Control}{clay}
\qrow[2D pose]{Marionette}{marionette}
\qrow[skeleton]{Fun-5B}{fun5b_pose_gt}
\qrow[depth]{Fun-5B}{fun5b_depth}
\qrow[G-buffer]{World Renderer}{worldrenderer_gbuf}
\qrow[3D proxies]{Fun-5B}{fun5b_proxy}
\qrow[teacher]{\ours}{ours_teacher}
\qrow[Stage~1]{\ours}{ours_stage1}
\qrow[Stage~2]{\ours}{ours_stage2}
\qrow{Ground truth}{gt}
\caption{\textbf{Comparison with all baselines} on two engine test clips, the clip of \cref{fig:qualitative} and that of \cref{fig:overview} (crops identical for all rows; the control type is given under each method).
Marionette loses the scene and the weapons within seconds.
World Renderer, which takes the appearance from the albedo channel, renders white figures without a scene from our G-buffers, whose albedo is constant (\cref{supp:baselines}).
Fun-5B-Control keeps the scene with all three controls but shifts the colors of the characters; with skeletons, which carry no props, it places the weapons in other positions than the ground truth (the rater names a missing weapon in $27\%$ of its ratings, against $2\%$ with depth or our proxies).
Our teacher and both students stay close to the ground truth.}
\label{fig:supp_comparison}
\end{figure*}

\paragraph{Contact events}
\cref{fig:supp_contact} shows consecutive frames around three contact events (\cref{supp:eval}).
\providecommand{\cstrip}[2]{\raisebox{-0.5\height}{\includegraphics[width=0.293\linewidth]{fig/img/ct_#1_#2.jpg}}}
\providecommand{\crow}[3][]{\rowlab[#1]{#2}\hspace{3pt}\cstrip{#3}{0}\hspace{4pt}\cstrip{#3}{1}\hspace{4pt}\cstrip{#3}{2}\\[1.5pt]}
\providecommand{\ctime}[1]{\makebox[\dimexpr0.293\linewidth*256/776\relax]{\scriptsize #1}}
\providecommand{\cgap}{\hspace{\dimexpr0.293\linewidth*4/776\relax}}
\providecommand{\cheads}{\ctime{$t_0{-}1$}\cgap\ctime{\textbf{contact} $t_0$}\cgap\ctime{$t_0{+}1$}}
\begin{figure*}[t]
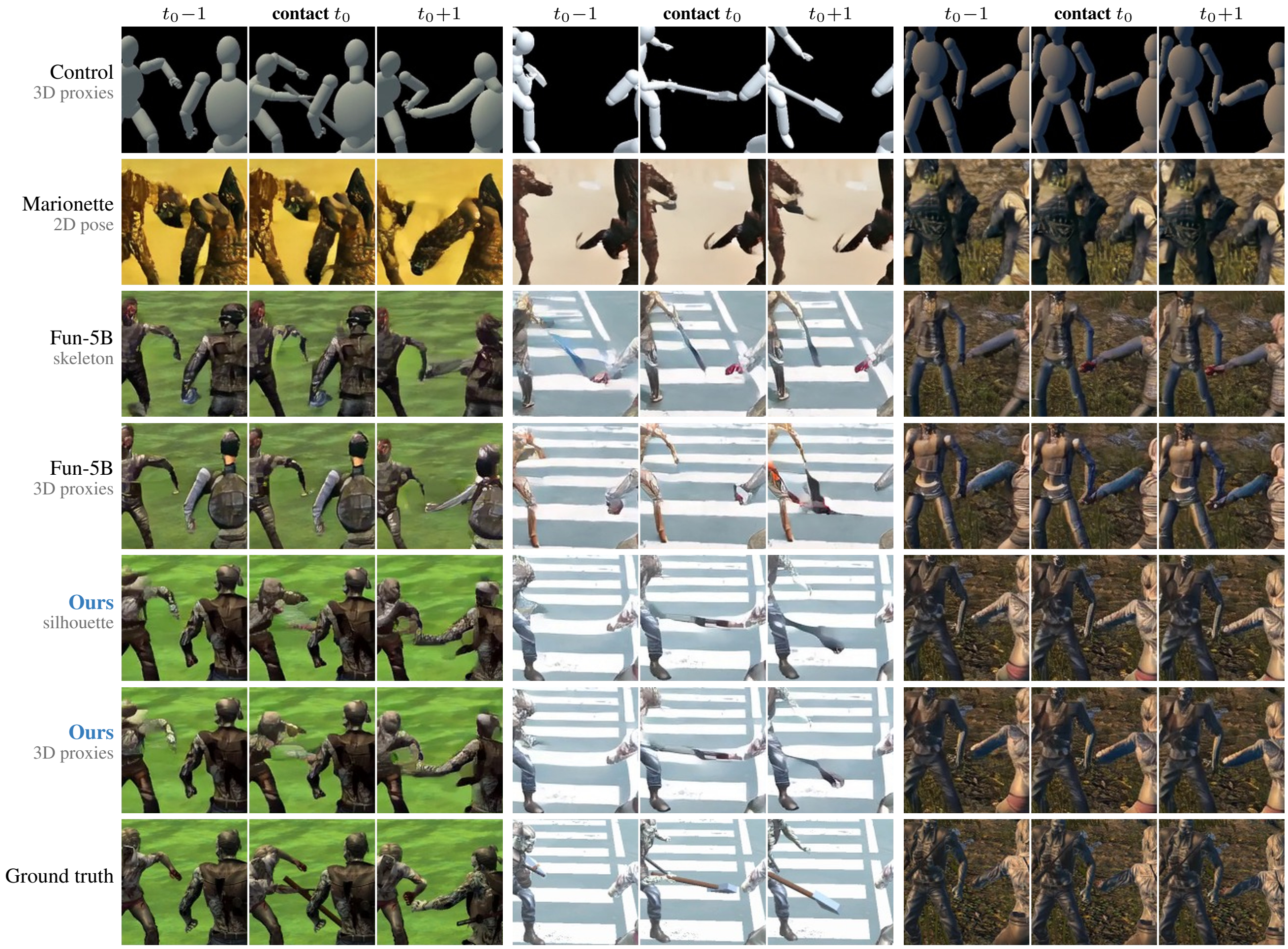

\centering
\rowlab{\mbox{}}\hspace{3pt}\cheads\hspace{4pt}\cheads\hspace{4pt}\cheads\\[1pt]
\crow[3D proxies]{Control}{ctl}
\crow[2D pose]{Marionette}{marionette}
\crow[skeleton]{Fun-5B}{fun5b_pose_gt}
\crow[3D proxies]{Fun-5B}{fun5b_proxy}
\crow[silhouette]{\ours}{abl_ctrl_silhouette}
\crow[3D proxies]{\ours}{ours_stage2}
\crow{Ground truth}{gtrow}
\caption{\textbf{Contact events.} Three consecutive frames ($1/16$\,s apart) around the frame $t_0$ at which the two silhouettes of the control merge, for three events chosen from the control alone, one per environment (windows of $128{\times}128$ pixels at the contact point, enlarged).
Left: an axe crosses between the characters and is visible for a single frame; our model renders only a faint trace of it, and the baselines none.
Middle: a swing that reaches the opponent; our model draws the weapon in the frame and at the place of the ground truth, darker and with less detail, while Fun-5B-Control draws weapon-like shapes in other colors and positions.
Right: an arm reaches the opponent; all methods except Marionette keep the pose, and Fun-5B-Control changes the colors of the arms.
A flat silhouette as control gives almost the same frames at these three events.}
\label{fig:supp_contact}
\end{figure*}

\paragraph{Frames of one-minute rollouts}
\cref{fig:drift} shows two one-minute rollouts (\cref{sec:long}; clip caption), with and without references.
\providecommand{\lrow}[3][]{\rowlab[#1]{#2}\hspace{3pt}%
  \raisebox{-0.5\height}{\includegraphics[width=0.445\linewidth]{fig/img/long_#3_0.jpg}}\hspace{4pt}%
  \raisebox{-0.5\height}{\includegraphics[width=0.445\linewidth]{fig/img/long_#3_1.jpg}}\\[1.5pt]}
\providecommand{\ltime}[1]{\makebox[\dimexpr0.445\linewidth*391/1576\relax]{\scriptsize #1}}
\providecommand{\lgap}{\hspace{\dimexpr0.445\linewidth*4/1576\relax}}
\providecommand{\lheads}{\ltime{$0.5$\,s}\lgap\ltime{$20$\,s}\lgap\ltime{$40$\,s}\lgap\ltime{$60$\,s}}
\begin{figure*}[t]
\centering
\rowlab{\mbox{}}\hspace{3pt}\lheads\hspace{4pt}\lheads\\[1pt]
\lrow[3D proxies]{Control}{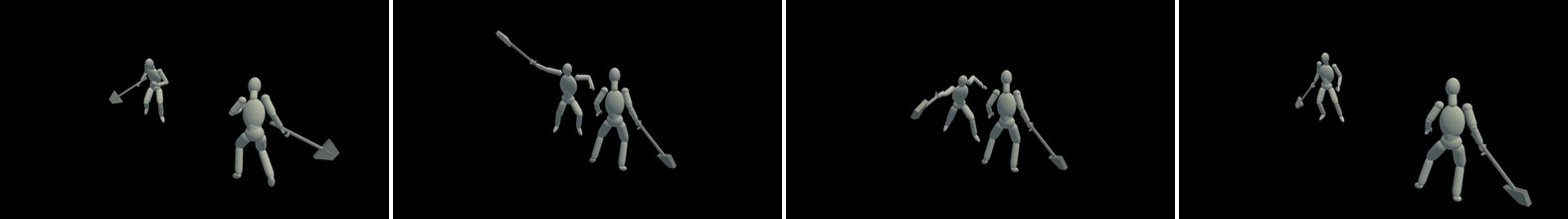}
\lrow{\ours}{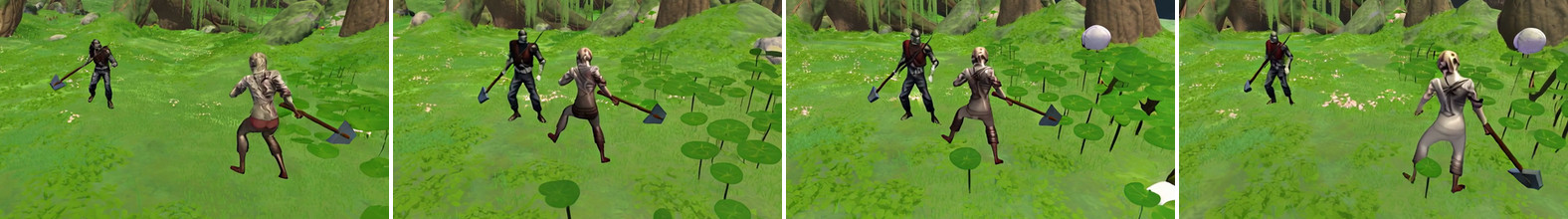}
\lrow[w/o references]{\ours}{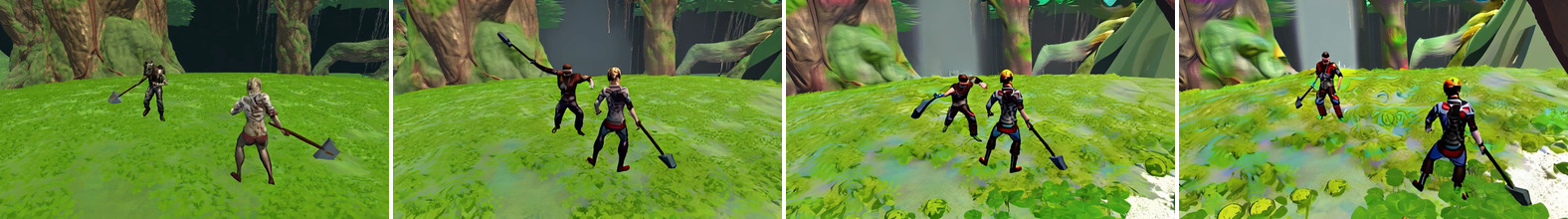}
\lrow{Ground truth}{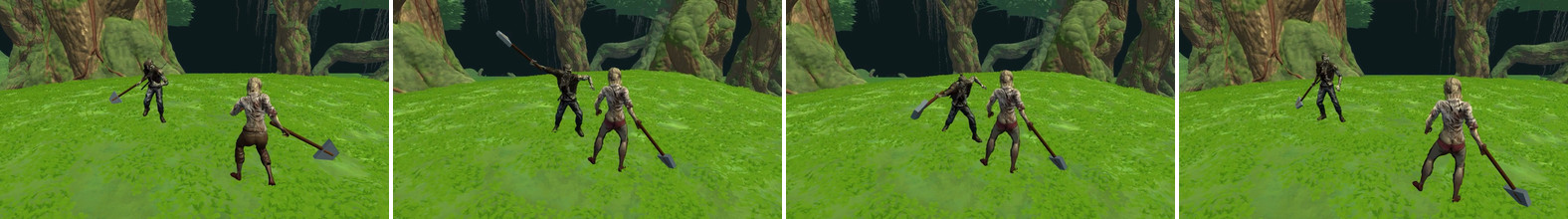}
\vspace{2pt}
\includegraphics[width=\linewidth]{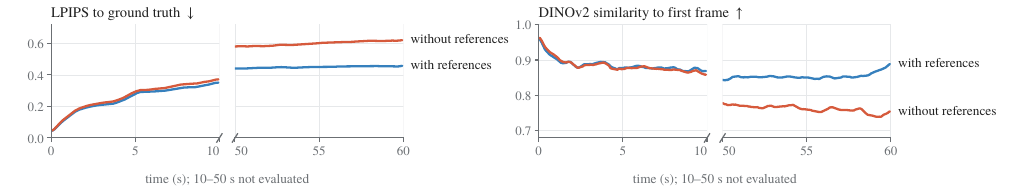}
\caption{\textbf{One-minute rollouts} of Stage~2 on two engine test clips of unseen characters, with the control played forward and backward for $60.6$\,s and generated in one rollout without resetting the cache (crops identical for all rows).
With the references in the cache, both characters keep their identities and the scene its layout for one minute, but details drift: objects that are not in the scene appear after $20$\,s and the clothing of one character lightens by $60$\,s (left clip).
Without references, the clothing of both characters changes within $20$\,s and the colors of the scene saturate by $40$\,s; the first frame has left the cache after $4.3$\,s.
Bottom: per-frame LPIPS to the ground truth and DINOv2 similarity to the first generated frame, averaged over the $19$ rollouts in the two evaluated windows (curves smoothed over $0.5$\,s).}
\label{fig:drift}
\end{figure*}

\paragraph{Inference-time ablations}
\cref{fig:supp_ablation} shows the appearance and control variants of \cref{tab:ablation} on two engine test clips.
With references only, the identities can be attached to the wrong proxies (right clip): identical white proxies do not tell which reference belongs to which character, and the first frame fixes the assignment.
\providecommand{\ablstrip}[1]{\raisebox{-0.5\height}{\includegraphics[width=\dimexpr(\linewidth-1.6cm-7pt)/2\relax]{fig/img/abl_#1.jpg}}}
\providecommand{\arow}[2]{\parbox[c]{1.6cm}{\raggedleft\scriptsize #1}\hspace{3pt}\ablstrip{#2_0}\hspace{3pt}\ablstrip{#2_1}\\[1.5pt]}
\providecommand{\ahead}[1]{\parbox[c]{1.6cm}{\mbox{}}\hspace{3pt}\makebox[\dimexpr\linewidth-1.6cm-4pt\relax]{\footnotesize #1}\\[2pt]}
\begin{figure*}[t]
\centering
\begin{minipage}[t]{0.493\linewidth}
\ahead{(a) Appearance inputs}
\arow{Control\\{\color{black!55}3D proxies}}{clay}
\arow{None}{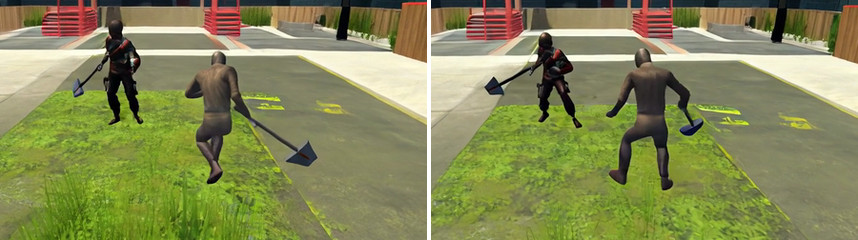}
\arow{References only}{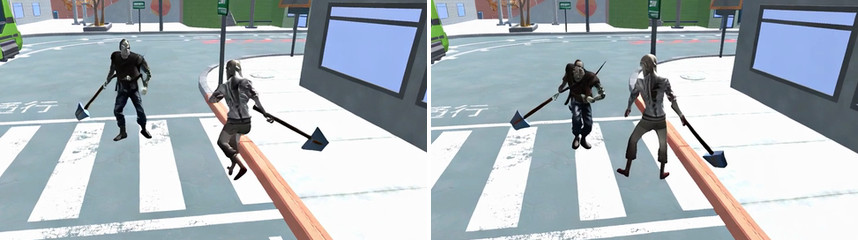}
\arow{First frame only}{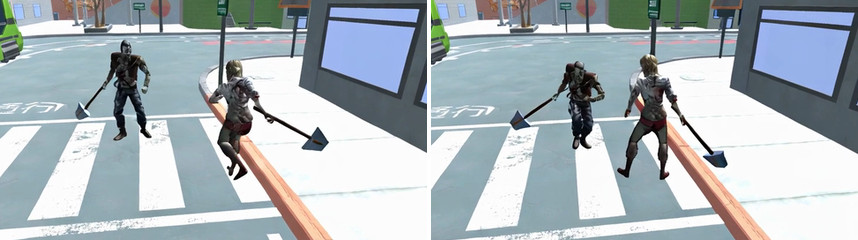}
\arow{First frame\\+ references\\{\bfseries\color{prGenLine}default}}{ours_stage2}
\end{minipage}\hfill
\begin{minipage}[t]{0.493\linewidth}
\ahead{(b) Control}
\arow{Ground truth}{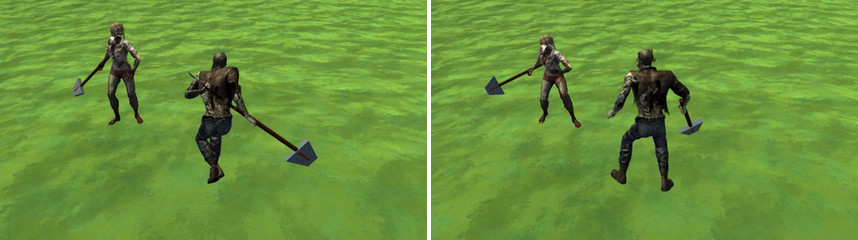}
\arow{Empty control}{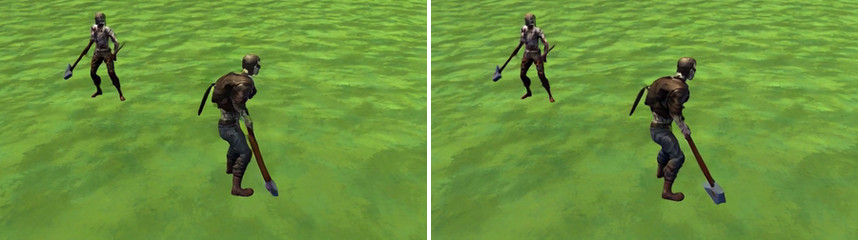}
\arow{Control of\\another clip}{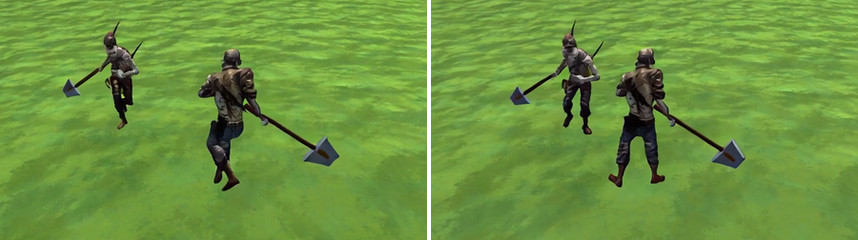}
\arow{Skeleton from\\ground truth}{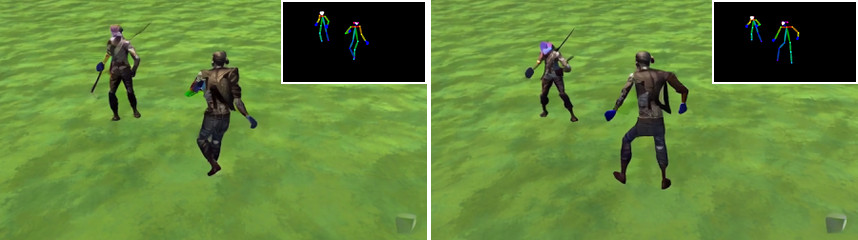}
\arow{Flat silhouette}{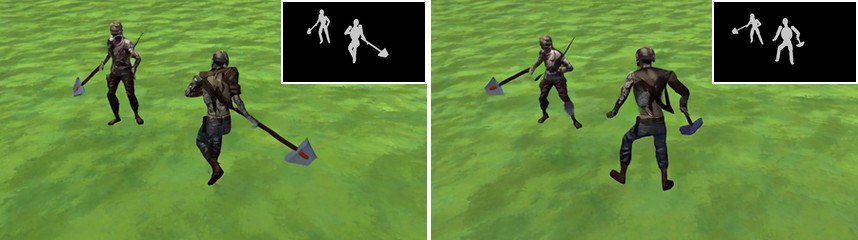}
\end{minipage}
\caption{\textbf{Inference-time ablations} of Stage~2 (\cref{tab:ablation}) on the two engine test clips of \cref{fig:supp_comparison} at $1.5$ and $4.5$\,s, without retraining (crops identical for all rows).
(a)~Without appearance inputs, the model invents the scene and the characters; references alone recover the scene but can attach the two identities to the wrong proxies (right clip), and the first frame resolves this.
(b)~With an empty control the characters barely move, and with the control of another clip they move differently from the ground truth. A skeleton estimated on the ground-truth frames (inset), which the model never saw in training, makes the weapons disappear and leaves colored traces at hands and heads; a flat silhouette (inset) gives almost the same frames at these two times (last row of (a)).}
\label{fig:supp_ablation}
\end{figure*}

\paragraph{External clips by dataset}
\cref{tab:supp_sources} splits the external results.
The teacher is the best model on MultiCamVideo and close to Fun-5B-Control on OmniWorld and MIND.
Stage~1 and Stage~2 each lose fidelity on every dataset, from $15.29$ to $12.05$\,dB on MultiCamVideo and from $17.28$ to $14.83$\,dB on OmniWorld, so the loss comes from both distillation stages and is not specific to one source.
Three observations narrow its cause. The teacher, which receives the same reconstructed proxies, is the best model on these clips, so the proxies and the domain alone do not explain it. Both conversions contribute ($1.5$\,dB from the teacher to Stage~1 and $1.3$\,dB from Stage~1 to Stage~2). And the loss is present from the first chunk and does not grow: on the $70$ clips with all $80$ frames, Stage~2 is $4.8$\,dB $[4.1, 5.5]$ below the teacher in frames $1$ to $8$ and $2.2$\,dB $[1.6, 2.7]$ in frames $69$ to $80$ (Stage~1: $2.6$ and $1.2$\,dB), so it is not an accumulation of errors over generated history; on engine renders the same pattern is weaker ($1.4$ and $0.7$\,dB).
In these clips the camera moves and the proxies cover only the human bodies; a causal model that has seen little control and history may anticipate the scene motion worse than the teacher, which sees the whole control, but we did not test this.

\begin{table}[h]
\centering
\small
\caption{\textbf{External clips by dataset.} PSNR / LPIPS. MiraData contributes one clip and is omitted. Marionette: $46$, $25$ and $8$ clips.}
\label{tab:supp_sources}
\setlength{\tabcolsep}{3pt}
\resizebox{\linewidth}{!}{%
\begin{tabular}{@{}lccc@{}}
\toprule
Method & MultiCamVideo ($70$) & OmniWorld ($28$) & MIND ($9$) \\
\midrule
Marionette & 13.01 / 0.595 & 15.00 / 0.569 & 12.16 / 0.600 \\
Fun-5B-Control, skeleton & 14.86 / 0.479 & 17.46 / 0.509 & 14.08 / 0.530 \\
Fun-5B-Control, depth & 14.12 / 0.513 & 17.44 / 0.507 & 13.89 / 0.541 \\
Fun-5B-Control, proxies & 14.30 / 0.514 & 17.36 / 0.515 & 13.88 / 0.541 \\
World Renderer & 9.06 / 0.882 & 12.05 / 0.806 & 9.73 / 0.816 \\
Ours, teacher & 15.29 / 0.437 & 17.28 / 0.509 & 13.21 / 0.533 \\
Ours, Stage~1 & 13.36 / 0.573 & 16.52 / 0.627 & 13.05 / 0.595 \\
Ours, Stage~2 & 12.05 / 0.610 & 14.83 / 0.627 & 12.43 / 0.574 \\
\quad first frame only & 11.14 / 0.641 & 14.38 / 0.636 & 12.04 / 0.596 \\
\bottomrule
\end{tabular}}
\end{table}

\section{Supplementary video}
\label{supp:videos}
The supplementary video, the ancillary file \texttt{PlayRender\_supplementary\_video.mp4} ($131$\,s, $1920{\times}1080$ pixels, H.264, no audio), shows clips of the paper in motion; \cref{tab:supp_video} lists its parts.
All clips have $864{\times}480$ pixels and $16$ frames per second. The video runs at $32$ frames per second and shows every frame of a clip twice, so that clips play in real time unless the table gives another speed; close-ups show the window of $128$ pixels of \cref{supp:contact} over the frames $t_0-10$ to $t_0+10$.
Clips are named by the first eight characters of their identifiers in our data.
The clips and events of the video are examples that we selected for display; all numbers of the paper are computed on the complete sets.
The four events of the fourth part are those of \cref{fig:qualitative}, which were chosen from the control alone; the events of the fifth and of the last part were chosen by us after looking at the outputs, as examples of a strike that Stage~2 renders and of the incomplete weapons that the direct check counts (\cref{tab:supp_human}).
On the engine test clips the video shows Stage~2 with the clip caption and on the new characters with the static caption (\cref{supp:eval}, text prompt).

\begin{table*}[t]
\centering
\small
\caption{\textbf{Parts of the supplementary video.} Time in minutes and seconds; the number after a clip is the contact frame $t_0$ of its event.}
\label{tab:supp_video}
\setlength{\tabcolsep}{5pt}
\begin{tabular}{@{}lp{8.0cm}p{5.7cm}l@{}}
\toprule
Time & Content & Clips & Speed \\
\midrule
0:04 & Control, output of Stage~2, first frame, references and ground truth for the three clips of \cref{fig:teaser} & \texttt{2e7ec0c8}, \texttt{20cda6d9}, \texttt{373ce72d} & $1\times$ \\
0:19 & Two further clips with new characters & \texttt{2b0a85f3}, \texttt{1d832247} & $1\times$ \\
0:24 & All baselines on three engine test clips, the first two those of \cref{fig:qualitative,fig:overview} & \texttt{e803cbef}, \texttt{230d85d7}, \texttt{4c4d0709} & $1\times$ \\
0:39 & Teachers of one budget with skeletons and with proxies at the four events of \cref{fig:qualitative} & \raggedright\texttt{1d7052b3} ($42$), \texttt{1d53ad0a} ($70$), \texttt{1ca396a6} ($35$), \texttt{230d85d7} ($71$) & $\tfrac14\times$ \\
1:00 & Fun-5B-Control (depth) and Stage~2 at three contact events & \raggedright\texttt{1e41fd64} ($70$), \texttt{2556be1e} ($34$), \texttt{230d85d7} ($71$) & $\tfrac14\times$ \\
1:16 & Teacher, Stage~1 and Stage~2 & \texttt{230d85d7} & $1\times$ \\
1:21 & Inference-time ablations: appearance inputs, then control & \texttt{230d85d7} & $1\times$ \\
1:31 & One-minute rollout with and without references, left clip of \cref{fig:drift} & \texttt{1c814956} & $4\times$ \\
1:46 & Two of the external clips of \cref{fig:results} & MultiCamVideo, MiraData & $1\times$ \\
1:56 & Limitation: incomplete weapons at two contact events & \texttt{b1069245} ($41$), \texttt{3e19b345} ($70$) & $\tfrac14\times$ \\
\bottomrule
\end{tabular}
\end{table*}

\end{document}